\pdfoutput=1
\documentclass[twocolumn,letterpaper]{main}
\paperstyle{fancy}
\papercolor{blue}
\usepackage[scaled=0.9]{helvet}
\makeatletter
\let\arxiv@xfloat\@xfloat
\def\@xfloat#1[#2]{\arxiv@xfloat#1[#2]\def\@tempa{table}\ifx\@captype\@tempa\small\fi}
\makeatother

\titleformat{\section}{\Large\sffamily\bfseries\color{accentcolor}\raggedright}{\thesection}{0.6em}{}
\titleformat{\subsection}{\large\sffamily\bfseries\color{accentcolor}\raggedright}{\thesubsection}{0.5em}{}
\titleformat{\subsubsection}{\normalsize\sffamily\bfseries\color{accentcolor}\raggedright}{\thesubsubsection}{0.4em}{}
\titleformat{\paragraph}[runin]{\normalfont\normalsize\sffamily\bfseries}{}{0pt}{}
\titlespacing*{\paragraph}{0pt}{1.4ex plus .3ex minus .1ex}{0.8em}

\usepackage{tikz}
\usepackage{placeins}  %
\definecolor{fagent}{HTML}{3D85C6}
\definecolor{fother}{HTML}{CC4125}
\definecolor{fscene}{HTML}{7F7F7F}
\definecolor{fappear}{HTML}{B8860B}
\DeclareRobustCommand{\fbadge}[2]{\tikz[baseline=(fb.base)]\node[circle,fill=#1,text=white,inner sep=0pt,minimum size=1.0em,font=\sffamily\bfseries\scriptsize](fb){#2};}
\providecommand{\cone}{\fbadge{fagent}{1}}
\providecommand{\ctwo}{\fbadge{fother}{2}}
\providecommand{\cthree}{\fbadge{fscene}{3}}
\providecommand{\cfour}{\fbadge{fappear}{4}}

\usepackage{array}
\usetikzlibrary{arrows.meta,calc}

\title{Learning Agent States for Latent Action Modeling}
\author[1]{Shaoan Xie}
\author[2]{Fan Feng}
\author[3]{Peiyuan Zhu}
\author[1]{Peter Spirtes}
\author[1,3]{Kun Zhang}
\affiliation[1]{Carnegie Mellon University}
\affiliation[2]{University of California, San Diego}
\affiliation[3]{Mohamed bin Zayed University of Artificial Intelligence}

\abstract{%
Latent action models (LAMs) learn action-like latents from unlabeled video by coupling frame-pair inference with latent-conditioned prediction, enabling policy and world-model pretraining.
However, what the latent must explain depends on the representation space used for action inference and transition prediction.
When this space entangles the agent's evolving physical state with visual context, such as background and persistent agent appearance, inferred latents can encode information unrelated to the agent's actions.
To address this challenge, we propose Agent-State LAM, which jointly learns a compact agent-state representation and latent actions modeling its transitions.
The compact state aggregates dense visual features into a shared space for both action inference and forward prediction.
To guide what this state retains, we introduce spatially matched temporal discrimination (SMTD), which trains the state to distinguish its source frame from nearby frames through spatially flexible matching with frozen local visual features.
Rather than comparing global frame representations, SMTD scores each frame by emphasizing its strongest local feature matches across spatial locations.
This scoring allows background and persistent agent appearance to remain matched across frames despite displacement, making these shared cues less informative for identifying the source frame and encouraging the representation to capture the agent's evolving physical state.
Extensive experiments on distracting control tasks and six real robot video datasets demonstrate improved action recovery, downstream control, and world-model prediction.
}
\project{https://agent-state-lam.github.io}
\code{https://github.com/Mid-Push/Agent-State-LAM}

\newcommand{\maketitlesupplementary}{%
  \newpage
  \twocolumn[{\raggedright\LARGE\sffamily\bfseries\color{titlefg}Supplementary Material\par}\vspace{1.0em}]}

\makeatletter

\newcommand{\suppfloatpages}{%
  \setlength{\@dblfptop}{0pt}\setlength{\@dblfpsep}{14pt plus 2pt minus 2pt}\setlength{\@dblfpbot}{0pt plus 1fil}%
  \setlength{\@fptop}{0pt}\setlength{\@fpsep}{14pt plus 2pt minus 2pt}\setlength{\@fpbot}{0pt plus 1fil}%
}
\def\supp@toc{toc}
\newcommand{\suppcontents}{%
  \let\supp@addtocontents\addtocontents
  \renewcommand{\addtocontents}[2]{%
    \edef\supp@ext{##1}%
    \ifx\supp@ext\supp@toc \expandafter\@firstoftwo \else \expandafter\@secondoftwo \fi
    {\supp@addtocontents{apptoc}}{\supp@addtocontents{##1}}{##2}}}
\makeatother

\begin{document}
\maketitle
\section{Introduction}
\label{sec:intro}

Vision-language-action models (VLAs)~\citep{kim2024openvla,intelligence2025pi_}, world models~\citep{zhou2024dino,gao2026dreamdojo}, and world action models (WAMs)~\citep{ye2026world,bi2026motus} show promise for learning to act, predict, and plan in the physical world.
Their action-conditioned components typically require paired observations and actions, which are costly to collect at scale.
Latent action models (LAMs) offer a way to exploit abundant unlabeled video by coupling inverse and forward dynamics: the inverse model infers a latent action from a frame pair, and the forward model uses this latent together with the earlier frame to predict the later frame or its features~\citep{schmidt2024learning,bruce2024genie,nikulin2025latent}.
To discourage shortcuts such as encoding the target frame, these models often constrain latent capacity through vector quantization~\citep{ye2025latent,schmidt2024learning,bruce2024genie} or KL regularization~\citep{gao2025adaworld,gao2026dreamdojo}.
The inferred latents then serve as targets for policy pretraining~\citep{ye2025latent,bu2025univla} or as conditioning inputs for world models~\citep{bruce2024genie,gao2025adaworld}.

\begin{figure}[!t]
    \centering
    \includegraphics[width=0.86\linewidth]{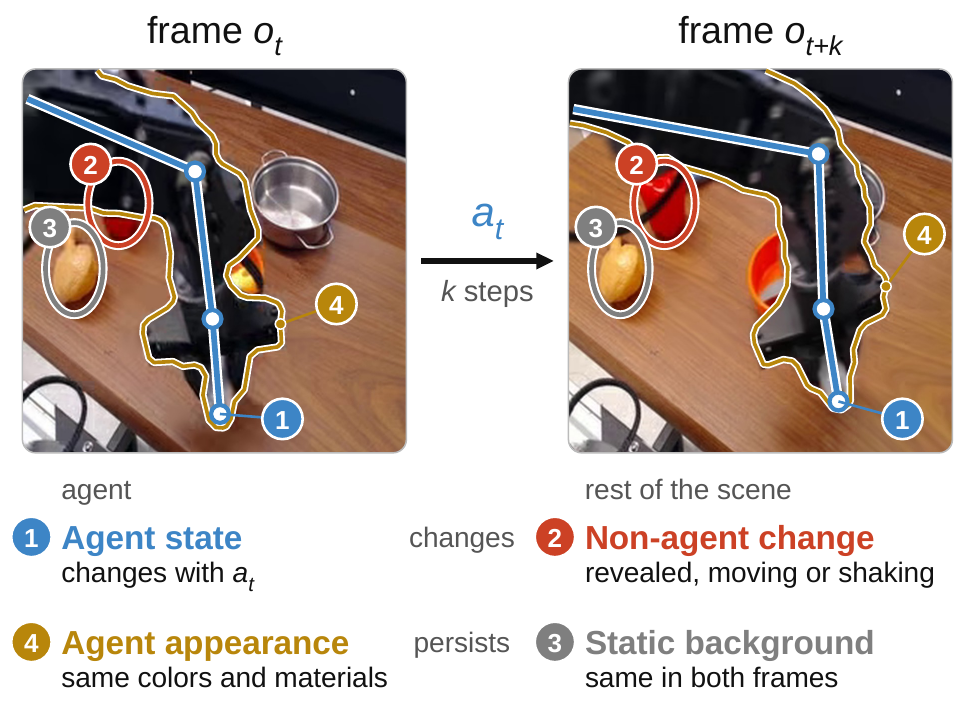}
    \caption{\textbf{Visual features used to infer latent actions can entangle the agent's physical state~\cone{} with visual context~\ctwo--\cfour{}.}
    Between the two video frames, the robot's physical state changes (\cone, illustrated by the blue pose skeleton), while its surface colors and materials persist (\cfour, outlined in gold).
    These are distinct factors of the same agent.
    The arm's motion also reveals previously occluded scene content (\ctwo), while other background content remains unchanged (\cthree).
    Such entanglement can obscure changes in the agent's physical state and introduce non-agent change into latent action inference.}
    \label{fig:visual_context}
\end{figure}

However, constraining latent capacity limits how much information the latent can encode, not which visual factors it captures.
The prediction objective instead encourages the latent to provide information about the later frame or its features that is not available from the earlier frame alone~\citep{lee2026latent,zhang2026latent}.
Consequently, the latent may encode any visual changes that help predict the target, including those unrelated to the agent's actions.
\textbf{What the latent captures therefore depends on the representation space, whether pixels or visual features, used for action inference and forward prediction}.
Figure~\ref{fig:visual_context} illustrates how this space can entangle the agent's evolving physical state~\cone{}, such as a robot arm's pose, with visual context.
Non-agent change~\ctwo{} can introduce irrelevant transition information into the latent, while static background~\cthree{} and persistent agent appearance~\cfour{} can obscure the physical-state changes needed to infer the agent's action.
We demonstrate these effects through controlled analyses on the Distracting Control Suite (DCS)~\citep{stone2021distracting}, which features natural-video backgrounds (Section~\ref{cs:natural}).
For LAPO~\citep{schmidt2024learning}, which predicts pixels, a linear probe recovers the spatial pattern of background change from its latent (up to $79\%$ of the pattern's variance).
For LAOM~\citep{nikulin2025latent}, which predicts learned features, replacing only the background while the agent's state is held fixed changes the features by up to $38\%$ (mean squared change, relative to randomly paired frames), showing that they remain entangled with the background.
Consistent with this background dependence, LAOM's nearest-neighbor matches fall within the $1\%$ closest poses for only $40\%$ of queries in cross-background pose retrieval (Table~\ref{tab:dcs-change}).

Such entanglement can persist even with additional guidance for latent action learning.
Spatial selection reduces background interference~\ctwo--\cthree{}: MaskLAM~\citep{fechner2026segment} restricts reconstruction to segmented agent pixels.
However, selecting agent content does not necessarily separate physical state from persistent agent appearance~\cfour{}: recoloring only the agent changes MaskLAM's representation by $9\%$ on average (up to $20\%$ on cheetah; Section~\ref{cs:natural}), and only $49\%$ of its queries meet the cross-background pose-retrieval criterion (Table~\ref{tab:dcs-change}).
Pretrained visual encoders may offer more robust representations~\citep{lee2026latent}, but they do not necessarily separate agent state from visual context.
Seq-$\Delta$REPA~\citep{jiang2026olaf} aligns latent actions with temporal changes in frozen pretrained video features, which can still reflect background dynamics rather than the agent's physical state.
Indeed, adding Seq-$\Delta$REPA's alignment objective to LAOM does not improve cross-background pose retrieval, with the success rate falling from $40\%$ to $37\%$ (Table~\ref{tab:dcs-change}).

These findings motivate us to explicitly shape the representation shared by action inference and forward prediction.
Building on feature-space LAMs~\citep{nikulin2025latent,bu2025univla}, we introduce \textbf{Agent-State LAM}, which jointly learns a compact agent-state representation and latent actions that model its transitions.
To limit the influence of visual context, we use a learned aggregation that maps dense visual features into this shared state, which provides the inputs to action inference and the targets for latent-conditioned prediction.
Compactness alone, however, does not determine which visual factors the state retains: the prediction objective can be satisfied even when the state omits physical-state distinctions needed to infer the agent's actions.

To guide what the state retains, we introduce \textbf{spatially matched temporal discrimination (SMTD)}, which trains each state to distinguish its source frame from nearby frames of the same video.
Rather than comparing global frame representations, SMTD scores each candidate frame through spatially flexible matching between the learned state and frozen local visual features.
This scoring allows shared background content~\ctwo--\cthree{} and persistent agent appearance~\cfour{} to remain matched across frames even when displaced, making these shared cues less informative for identifying the source frame.
SMTD thus encourages the state to capture temporally distinctive information about the agent's evolving physical state~\cone{}.

Our main contributions are summarized as follows.
\begin{itemize}
    \item \textbf{Agent-state latent action modeling.} We introduce Agent-State LAM, a feature-space LAM that learns a compact agent state jointly with latent actions to model its transitions. This state provides a shared representation for action inference and forward prediction, while SMTD guides what it retains by training each state to distinguish its source frame from nearby frames through spatially flexible matching with frozen local visual features, without action labels, masks, or optical-flow targets.
    \item \textbf{Controlled analysis and downstream validation.} We systematically analyze how non-agent changes and persistent visual context affect latent actions and learned representations, including those guided by masks or pretrained visual features. We demonstrate improvements in action recovery, robustness to visual context, and downstream control on DCS, with further validation on six real robot video datasets and world-model prediction driven by decoded latent actions.
\end{itemize}

\section{Related work}
\label{sec:related}
\begin{figure*}[t]
    \centering
    \includegraphics[width=\textwidth]{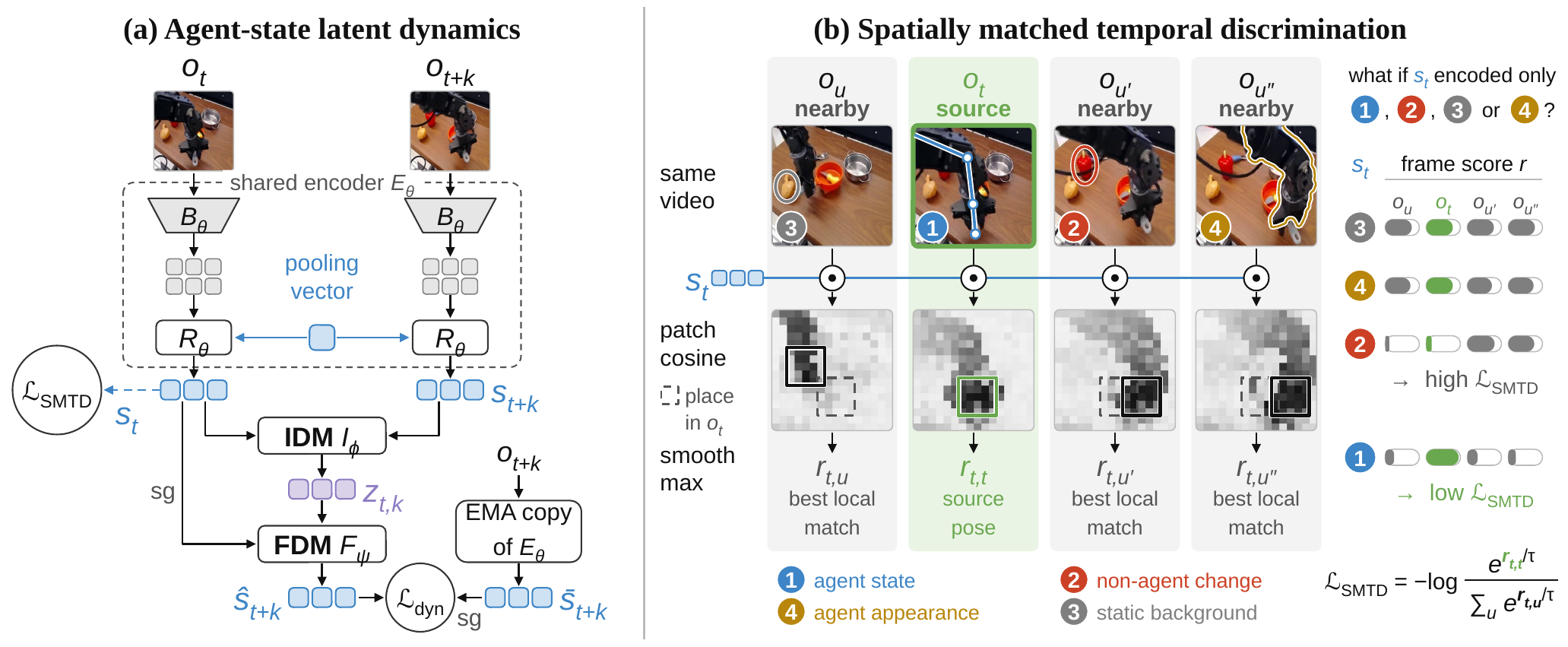}
    \caption{\textbf{Agent-State LAM.}
    (a)~The encoder $E_\theta$ (backbone $B_\theta$, attention pooling $R_\theta$) maps each frame to a compact agent state; the inverse model $I_\phi$ infers $z_{t,k}$ from two states, and the forward model $F_\psi$ predicts $\hat{s}_{t+k}$ against the EMA target $\bar{s}_{t+k}$ (sg: stop-gradient).
    (b)~SMTD on frames of one robot video: the projected state is compared with every frozen patch feature of a frame (gray maps, darker is more similar; solid box: best local match; dashed: its place in $o_t$), and the smooth maximum over patches gives the frame score $r_{t,u}$.
    Right: illustrative scores if $s_t$ encoded only one factor. Static background~\cthree{} and persistent agent appearance~\cfour{}, matched in every frame, and the pepper~\ctwo{}, hidden by the arm in $o_t$ and matched only in $o_{u'}$ and $o_{u''}$, give the source no higher score and a high $\mathcal{L}_{\mathrm{SMTD}}$; the agent's state~\cone{}, matched best at the source pose, gives a low loss.
    The maps use the gripper's frozen feature in $o_t$ in place of the projected state.}
    \label{fig:method}
\end{figure*}

\paragraph{Latent action models.}
LAM designs differ in the latent parameterization (continuous latents~\citep{liang2025clam,garrido2026learning} or structural constraints~\citep{wei2026learning,chen2026learning}) and in the prediction target (pixels~\citep{schmidt2024learning,bruce2024genie}, frozen DINO features~\citep{bu2025univla}, or learned features~\citep{nikulin2025latent}); broader design choices are studied by \citet{bu2026matters}.
Analyses of linear LAMs show that reconstruction objectives can allow exogenous variation to enter inferred latents~\citep{zhang2026latent,lee2026latent}, and highlight the importance of representations that are less sensitive to exogenous changes~\citep{lee2026latent}.
Motivated by this theoretical insight, we jointly learn latent actions and a compact agent state for action inference and forward prediction, with SMTD guiding what it retains.

\paragraph{Guidance for latent actions and visual representation learning.}
Action supervision in LAOM-Labels~\citep{nikulin2025latent}, task language in UniVLA~\citep{bu2025univla}, and prompt-conditioned VLM prediction targets~\citep{nikulin2026vision} guide learning toward action-relevant information.
Spatial selection focuses learning on relevant regions through object slots selected by action-supervised probes~\citep{klepach2026object} or segmentation masks in MaskLAM~\citep{fechner2026segment}.
LAOF~\citep{bu2026laof} uses optical-flow targets to emphasize motion, while factored LAMs~\citep{wang2026factored} assign separate latent actions to different scene factors.

More broadly, joint-embedding predictive architectures (JEPAs) learn representations through feature prediction~\citep{lecun2022path,assran2023self,bardes2024revisiting,assran2025v,maes2026leworldmodel}, using momentum target encoders or variance and distribution regularizers to stabilize learning~\citep{grill2020bootstrap,bardes2021vicreg,balestriero2025lejepa}.
These mechanisms help prevent representational collapse but do not determine which visual information is retained.
Temporal comparisons have also been used in time-contrastive networks~\citep{sermanet2018time,dwibedi2018learning}, Premier-TACO~\citep{zheng2024premier}, and CRTR~\citep{ziarko2026contrastive}, which uses within-trajectory negatives to discourage static-context shortcuts.
Within LAMs, CLAP aligns visual latents with robot-action representations~\citep{zhang2026clap}; ConLA uses action-category labels to contrast motion embeddings and temporal reversal to align appearance embeddings~\citep{dai2026conla}; and CoMo contrasts motion latents from temporally shifted and reversed transitions~\citep{yang2026learning}.
Seq-$\Delta$REPA~\citep{jiang2026olaf} further aligns latent actions with temporal changes in frozen video features.
In contrast, SMTD uses nearby-frame discrimination to shape the shared per-frame state, with spatially flexible matching against frozen local visual features.

\section{Method}
\label{sec:method}

Agent-State LAM jointly trains a state encoder and inverse and forward dynamics models on unlabeled video.
Both dynamics models operate on a compact agent-state representation, whose content is shaped by latent-conditioned prediction and SMTD (Figure~\ref{fig:method}).

\subsection{Preliminaries: feature-space latent dynamics}
\label{sec:method:dynamics}

Feature-space LAMs infer latent actions from visual representations and use them to predict future representations~\citep{nikulin2025latent,bu2025univla}.
Given an unlabeled video $\mathcal V=(o_0,\ldots,o_T)$, an encoder $E_\theta$ maps each input $x_t$, either a frame $o_t$ or a short observation history ending at $t$, to a representation $s_t=E_\theta(x_t)\in\mathbb R^d$.
The inverse dynamics model (IDM) $I_\phi$ infers a latent action from two representations $k$ steps apart, and the forward dynamics model (FDM) $F_\psi$ uses it to predict the future representation:
\begin{equation}
    \begin{aligned}
    z_{t,k} &= Q\!\left(I_\phi(s_t,s_{t+k})\right),\\
    \hat{s}_{t+k} &= F_\psi(s_t,z_{t,k}).
    \end{aligned}
    \label{eq:method:latent-dynamics}
\end{equation}
Here, $Q$ is the identity for continuous latents or a quantizer for discrete codes.
The prediction is trained to match a target representation $\bar{s}_{t+k}$ of the future observation, provided by a frozen visual encoder~\citep{bu2025univla} or an exponential moving average (EMA) copy of the learned encoder~\citep{grill2020bootstrap,nikulin2025latent}:
\begin{equation}
    \mathcal L_{\mathrm{dyn}}
    =
    \mathbb E_{\mathcal V,t,k}
    \left[
        \frac{1}{d}
        \left\|
        \hat{s}_{t+k}
        -\operatorname{sg}(\bar{s}_{t+k})
        \right\|_2^2
    \right].
    \label{eq:method:dyn}
\end{equation}
The content of $s_t$ therefore determines both the information available for latent-action inference and the changes that the latent must help predict.
However, prediction alone does not ensure that $s_t$ preserves the agent's physical-state distinctions rather than irrelevant visual context, motivating the agent-state representation introduced next.

\subsection{Agent-state space}
\label{sec:method:state}
We seek a representation that preserves the agent's evolving physical state~\cone{} while limiting dependence on visual context~\ctwo--\cfour{} (Figure~\ref{fig:visual_context}).
Feature-space LAMs such as LAOM~\citep{nikulin2025latent} use dense, high-dimensional feature maps, leaving substantial capacity to retain background and persistent agent appearance information alongside the agent's physical state.
This context can also influence the inferred latent actions: background motion accounts for $40\%$ of the combined response of LAOM's latent to agent- and background-motion interventions, averaged across the four tasks (Table~\ref{tab:dcs-change}).
These findings motivate restricting the capacity of the representation shared by action inference and prediction, limiting how much visual context it can retain.
\paragraph{Compact state by learned aggregation.}
To this end, we replace the dense feature map with a compact vector $s_t$, obtained through learned aggregation of spatial features.
Specifically, the encoder of Section~\ref{sec:method:dynamics} takes the form $E_\theta=R_\theta\circ B_\theta$: a visual backbone $B_\theta$ followed by an attention-pooling map $R_\theta$~\citep{jaegle2021perceiver,alayrac2022flamingo}.
The pooling map aggregates $N$ spatial features of width $c$ into a state $s_t\in\mathbb{R}^{d}$, with $d\ll Nc$ (Table~\ref{cs:tab:settings}).
We learn the aggregation jointly with the latent dynamics, using the compact states as both the inputs to action inference and the targets for prediction.
This shared compact-state design limits representation capacity and reduces action error by $67\%$ on DCS hopper, from $3.46$ with dense features to $1.14$ (Table~\ref{tab:ablation}).

Compactness alone, however, cannot ensure that the retained information is the physical agent state~\cone{}.
For example, a state encoding only static background can remain constant within a video and be copied forward with $\mathcal L_{\mathrm{dyn}}=0$.
This calls for a complementary learning signal that guides the compact representation toward the agent's evolving physical state.

\subsection{Spatially matched temporal discrimination}
\label{sec:method:smtd}

Temporal comparisons provide such a signal.
Static background~\cthree{} and persistent agent appearance~\cfour{} recur across nearby frames, the latter wherever the agent moves.
Non-agent change~\ctwo{} can also become persistent: for example, once the pepper in Figure~\ref{fig:visual_context} is revealed by the moving arm, it may remain visible in subsequent frames.
Likewise, moving distractors may retain their local appearance despite changing positions.
Matching visual content across spatial locations and contrasting nearby frames makes these shared cues less informative for identifying the source frame, favoring the agent's evolving physical state~\cone{} when its changes are more distinctive.
We therefore introduce spatially matched temporal discrimination (SMTD), which trains each compact state to identify its source frame among nearby frames of the same \mbox{video (Figure~\ref{fig:method}b).}

\begin{table*}[t]
\centering
\setlength{\tabcolsep}{3pt}
\caption{\textbf{Action recovery and control in DCS.} Stage~1: test error of a linear probe mapping inferred latent actions to ground-truth actions. Stage~2: normalized return of a policy cloned on latent actions, with an action decoder trained on all available action labels. Stage~3: return of the same policy with a decoder trained on only $32$ labeled trajectories, following~\cite{nikulin2025latent}.}
\label{tab:dcs}
\vspace{3pt}
\begin{tabular*}{\textwidth}{@{\extracolsep{\fill}} l cccc cccc cccc}
\toprule
 & \multicolumn{4}{c}{Stage 1: latent actions} & \multicolumn{4}{c}{Stage 2: behavior cloning} & \multicolumn{4}{c}{Stage 3: action decoding} \\
 & \multicolumn{4}{c}{\textit{action error} $\downarrow$} & \multicolumn{4}{c}{\textit{return, all labels} $\uparrow$} & \multicolumn{4}{c}{\textit{return, $32$ trajectories} $\uparrow$} \\
\cmidrule(lr){2-5} \cmidrule(lr){6-9} \cmidrule(lr){10-13}
 & hop. & cheet. & walk. & hum. & hop. & cheet. & walk. & hum. & hop. & cheet. & walk. & hum. \\
\midrule
LAPO~\cite{schmidt2024learning} & 16.55 & 23.74 & 10.73 & 0.65 & 0.04 & 0.37 & 0.28 & 0.13 & 0.07 & 0.29 & 0.04 & 0.01 \\  %
MaskLAM~\cite{fechner2026segment} & 2.41 & 2.02 & 5.23 & 0.22 & 0.35 & 0.71 & 0.79 & 0.20 & 0.15 & 0.55 & 0.09 & 0.03 \\  %
\midrule
LAOM~\cite{nikulin2025latent} & 2.70 & 2.47 & 5.49 & 0.15 & 0.28 & 0.40 & 0.49 & 0.19 & 0.11 & 0.44 & 0.12 & 0.03 \\  %
\quad + Seq-$\Delta$REPA~\cite{jiang2026olaf} & 3.16 & 2.45 & 7.10 & 0.14 & 0.20 & 0.37 & 0.33 & 0.17 & 0.08 & 0.42 & 0.06 & 0.03 \\  %
\quad + optical-flow target~\cite{bu2026laof} & 2.35 & 2.32 & 3.06 & 0.14 & 0.30 & 0.56 & 0.65 & 0.06 & 0.15 & 0.49 & 0.17 & 0.02 \\  %
\quad + CoMo~\cite{yang2026learning} & 2.12 & 2.07 & 4.18 & 0.14 & 0.21 & 0.52 & 0.56 & 0.08 & 0.14 & 0.54 & 0.06 & 0.04 \\  %
\midrule
\textbf{Agent-State LAM} & \textbf{1.14} & \textbf{1.01} & \textbf{2.37} & \textbf{0.11} & \textbf{0.53} & \textbf{0.90} & \textbf{0.81} & \textbf{0.22} & \textbf{0.37} & \textbf{0.67} & \textbf{0.36} & \textbf{0.06} \\  %
\midrule
\textit{LAOM + $32$k labels} & \textit{1.04} & \textit{0.81} & \textit{2.02} & \textit{0.12} & \textit{0.43} & \textit{0.68} & \textit{0.62} & \textit{0.18} & \textit{0.27} & \textit{0.52} & \textit{0.26} & \textit{0.04} \\  %
\textit{LAOM + $128$k labels} & \textit{0.78} & \textit{0.67} & \textit{1.48} & \textit{0.11} & \textit{0.54} & \textit{0.72} & \textit{0.85} & \textit{0.17} & \textit{0.36} & \textit{0.66} & \textit{0.55} & \textit{0.05} \\  %
\bottomrule
\end{tabular*}
\end{table*}

\paragraph{Spatially flexible frame scoring.}
SMTD first scores each candidate frame against the learned state, then uses these scores for temporal discrimination.
To reduce sensitivity to the spatial displacement of shared visual content, we allow the state to match local features across all locations in each candidate frame.
For each candidate frame $o_u$, a frozen visual encoder $D$ provides $N_p$ patch features $f_{u,p}$.
We project the compact state $s_t$ into the same feature space, $q_t=q(s_t)$.
The cosine similarities $\cos(q_t,f_{u,p})$ measure how strongly each patch matches the state.
We aggregate these similarities through a smooth maximum over patch locations, with temperature $\tau_p>0$, to obtain the frame score:
\begin{equation}
    r_{t,u}
      =\tau_p\log\!\sum_{p=1}^{N_p}
        \exp\!\left(\frac{\cos(q_t,f_{u,p})}{\tau_p}\right).
    \label{eq:method:score}
\end{equation}
The smooth maximum emphasizes the strongest local matches without requiring them to occur at fixed spatial locations.
Figure~\ref{fig:method}b illustrates how these matches can follow the displaced arm across nearby frames (solid boxes), rather than remaining at its source-frame location (dashed boxes).
This spatial flexibility is also reflected in our DCS hopper analysis: matching across locations roughly halves the measured change in distractor patches, leaving it below the agent's change in nine of ten trajectories (Section~\ref{cs:smtd}).

\paragraph{Temporal contrast.}
We use the spatially matched scores $r_{t,u}$ to distinguish each state's source frame from nearby frames.
For video $\mathcal V$, let $\mathcal C_{\mathcal V}(t)$ contain the source time $t$ as the positive, and negative times sampled from a window of nearby frames.
With temperature $\tau>0$, we minimize
\begin{equation}
    \mathcal L_{\mathrm{SMTD}}
      =-\mathbb E_{\mathcal V,t}\log
      \frac{\exp(r_{t,t}/\tau)}
      {\sum_{u\in\mathcal C_{\mathcal V}(t)}
        \exp(r_{t,u}/\tau)}.
    \label{eq:method:smtd}
\end{equation}
Shared visual context~\ctwo--\cfour{} can also be matched in nearby frames, at the same or a shifted location, giving the source no higher score.
The agent's physical state~\cone{}, in contrast, changes from frame to frame and can match the source pose more strongly than nearby poses (Figure~\ref{fig:method}b, right).
Together with limited state capacity, this contrast encourages the representation to retain source-specific agent-state information rather than shared visual context.

\subsection{Joint optimization}
\label{sec:method:training}

We optimize prediction and SMTD jointly:
\begin{equation}
    \mathcal{L}=w_{\mathrm{dyn}}\mathcal{L}_{\mathrm{dyn}}+\mathcal{L}_{\mathrm{aux}}+\lambda\,\mathcal{L}_{\mathrm{SMTD}},
    \label{eq:method:objective}
\end{equation}
where $w_{\mathrm{dyn}}$ and $\lambda$ weight the prediction and SMTD objectives, and $\mathcal{L}_{\mathrm{aux}}$ is the quantization objective for discrete codes~\citep{bu2025univla} and zero for continuous latents.

\section{Experiments}
\label{sec:exp}

We evaluate whether learning an agent-state representation improves action recovery while reducing dependence on visual context.
DCS provides ground-truth actions and poses and allows controlled changes to background and agent appearance at a fixed physical state, enabling context analysis, downstream control evaluation, and ablations of the compact state and SMTD.
We then evaluate action recovery on six real robot video datasets and frozen world-model prediction using decoded latent actions.

\subsection{Experimental setup}
\label{sec:exp:setup}

\begin{table}[t]
\centering
\setlength{\tabcolsep}{2.4pt}
\caption{\textbf{Dependence on visual context}: pose retrieval and responses to controlled interventions, by factor (mean over the four DCS tasks). \emph{R@1} (\%) (\cone{}): cross-background pose retrieval, counted correct when the nearest-neighbor match is among the $1\%$ closest poses (chance $1\%$). \emph{Motion} (\ctwo{}): the background's share of the latent's response when background or agent motion is frozen. \emph{Static} (\cthree{}): normalized latent change when only the static background is swapped. \emph{Recolor} (\cfour{}): representation change when only the agent is recolored, relative to the distance between \mbox{randomly} paired frames.}
\label{tab:dcs-change}
\vspace{3pt}
\begin{tabular}{@{}l c cc c@{}}
\toprule
 & R@1 $\uparrow$ & share $\downarrow$ & \multicolumn{2}{c}{change $\downarrow$} \\
\cmidrule(lr){2-2} \cmidrule(lr){3-3} \cmidrule(l){4-5}
 & \cone{} & \ctwo{} & \cthree{} & \cfour{} \\
 & scene & motion & static & recolor \\
\midrule
LAPO & 18 & 0.76 & 0.69 & 0.02 \\
MaskLAM & 49 & 0.19 & 0.21 & 0.09 \\
\midrule
LAOM & 40 & 0.40 & 0.28 & 0.15 \\
\quad + Seq-$\Delta$REPA & 37 & 0.44 & 0.28 & 0.22 \\
\quad + optical-flow target & 46 & 0.44 & 0.26 & 0.12 \\
\quad + CoMo & 44 & 0.39 & 0.30 & 0.14 \\
\midrule
\textbf{Agent-State LAM} & \textbf{63} & \textbf{0.09} & \textbf{0.14} & \textbf{0.01} \\
\midrule
\textit{LAOM + $32$k labels} & \textit{55} & \textit{0.27} & \textit{0.17} & \textit{0.14} \\
\textit{LAOM + $128$k labels} & \textit{56} & \textit{0.25} & \textit{0.16} & \textit{0.13} \\
\bottomrule
\end{tabular}
\end{table}

\begin{table*}[t]
\makeatletter
\renewenvironment{table}[1][]{\begin{minipage}[t]{\columnwidth}\vspace{0pt}\def\@captype{table}}{\end{minipage}}%
\renewenvironment{figure}[1][]{\par\noindent\begin{minipage}[t]{\columnwidth}\vspace{0pt}\def\@captype{figure}}{\end{minipage}\par}%
\makeatother
\ifdefined\ablrhsbox\else\newsavebox{\ablrhsbox}\fi
\sbox{\ablrhsbox}{\providecommand{\halfcheck}{\tikz[baseline=(hc.base)]{\node[inner sep=0pt, outer sep=0pt] (hc) {\checkmark}; \draw[line width=0.45pt, line cap=round] ([xshift=0.36em, yshift=0.98ex]hc.south west) -- ++(0.36em,-0.50ex);}}
\begin{table}[t]
\centering
\setlength{\tabcolsep}{3.2pt}
\caption{\textbf{Ablations} (DCS hopper): Agent-State LAM with one change per row; metrics as in Tables~\ref{tab:dcs} and~\ref{tab:dcs-change}. Agent mask input: SAM~2 agent masks on the input in place of SMTD; without compact state: dense features throughout; dense IDM input: for the IDM only; frame-mean scoring: frame score against the mean teacher feature; state-to-state contrast: the model's own states of the same and nearby frames replace the frozen reference.}
\label{tab:ablation}
\vspace{3pt}
\begin{tabular}{@{}l cc c c@{}}
\toprule
 & compact & & action & pose \\
 & state & SMTD & error $\downarrow$ & R@1 $\uparrow$ \\
\midrule
LAOM (reference) & -- & -- & 2.70 & 32 \\  %
\textbf{Agent-State LAM} & \checkmark & \checkmark & \textbf{1.14} & \textbf{67} \\
\midrule
\multicolumn{5}{@{}l}{\textit{Model components}} \\
\hspace{0.6em}without SMTD & \checkmark & $\times$ & 3.32 & 10 \\
\hspace{0.6em}without compact state & $\times$ & \checkmark & 3.46 & 31 \\
\hspace{0.6em}agent mask input & \checkmark & $\times$ & 1.68 & 40 \\
\hspace{0.6em}dense IDM input & \halfcheck & \checkmark & 1.74 & 62 \\
\midrule
\multicolumn{5}{@{}l}{\textit{SMTD variants}} \\
\hspace{0.6em}positives only & \checkmark & \checkmark & 5.67 & 4 \\
\hspace{0.6em}frame-mean scoring & \checkmark & \checkmark & 2.48 & 26 \\
\hspace{0.6em}state-to-state contrast & \checkmark & \checkmark & 1.81 & 51 \\
\hspace{0.6em}DINOv2-S teacher & \checkmark & \checkmark & 1.22 & 62 \\  %
\bottomrule
\end{tabular}
\end{table}
}%
\begin{minipage}[t]{\columnwidth}\vspace{0pt}%
\begin{figure}[t]
\centering
\setlength{\tabcolsep}{0pt}%
\def\apw{\dimexpr(\linewidth-28pt)/5\relax}%
\def\aplab#1#2{\rotatebox{90}{\makebox[\apw]{\scriptsize\shortstack{#1\\[-1.5pt]#2}}}}%
\begin{tabular}{@{}c@{\hspace{2pt}}c@{\hspace{5pt}}c@{\hspace{1.5pt}}c@{\hspace{5pt}}c@{\hspace{1.5pt}}c@{}}
 & \scriptsize Source & \multicolumn{2}{c}{\scriptsize LAOM} & \multicolumn{2}{c}{\scriptsize Agent-State LAM} \\[1pt]
\aplab{Hopper}{Scene} & \includegraphics[width=\apw]{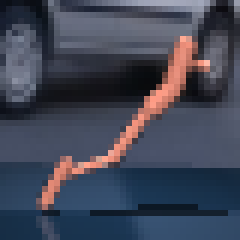} & \includegraphics[width=\apw]{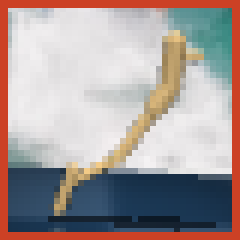} & \includegraphics[width=\apw]{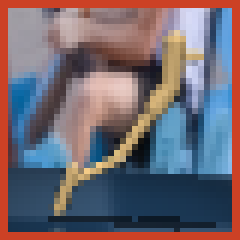} & \includegraphics[width=\apw]{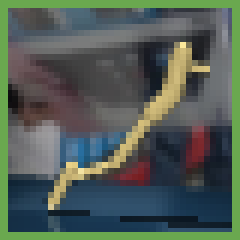} & \includegraphics[width=\apw]{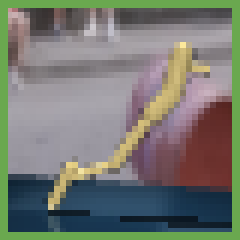} \\[1.5pt]
\aplab{Walker}{Color} & \includegraphics[width=\apw]{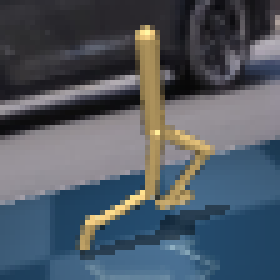} & \includegraphics[width=\apw]{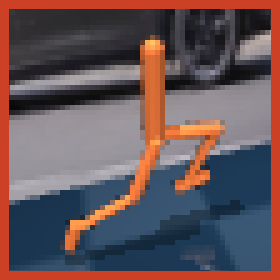} & \includegraphics[width=\apw]{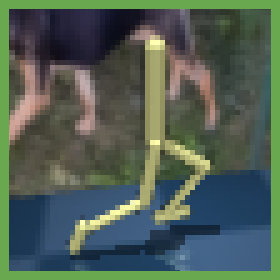} & \includegraphics[width=\apw]{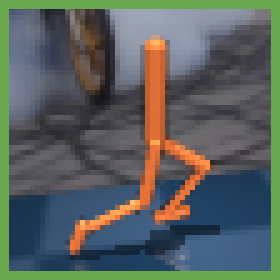} & \includegraphics[width=\apw]{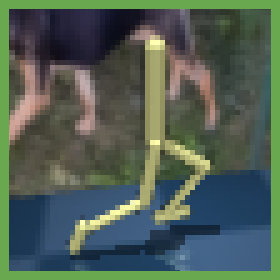} \\[1.5pt]
\aplab{Cheetah}{Thick} & \includegraphics[width=\apw]{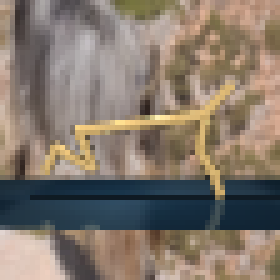} & \includegraphics[width=\apw]{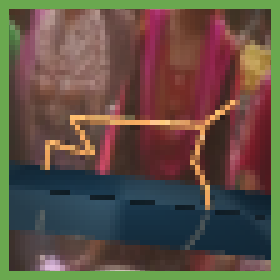} & \includegraphics[width=\apw]{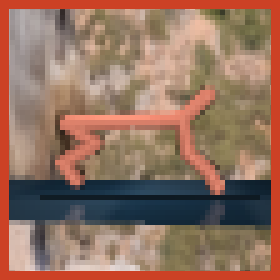} & \includegraphics[width=\apw]{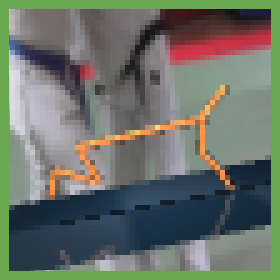} & \includegraphics[width=\apw]{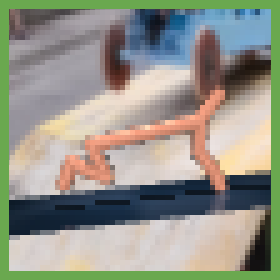} \\
\end{tabular}
\caption{\textbf{Pose retrieval across visual contexts} (DCS).
For each source frame, each model retrieves its nearest neighbor from each of two galleries of other trajectories, using cosine similarity between visual representations.
The galleries have new backgrounds~\ctwo--\cthree{}, with additional changes to agent color (walker) or limb thickness (cheetah)~\cfour{}.
Green borders indicate that the retrieved pose is among the $1\%$ closest gallery poses to the source; red borders indicate otherwise.
These examples illustrate that Agent-State LAM retrieves similar agent poses~\cone{} despite changes in background and agent appearance.}
\label{fig:dcs-change}
\end{figure}

\end{minipage}\hfill
\usebox{\ablrhsbox}%
\end{table*}

\paragraph{DCS.}
We build Agent-State LAM on LAOM's codebase~\cite{nikulin2025latent} and use its four DCS datasets~\cite{stone2021distracting}---hopper, cheetah, walker, and humanoid---which feature natural-video backgrounds and camera and color perturbations.
We compare against LAPO~\cite{schmidt2024learning}, LAOM, and MaskLAM~\cite{fechner2026segment}.
To compare auxiliary guidance objectives within a common architecture, we also incorporate Seq-$\Delta$REPA's alignment objective~\cite{jiang2026olaf}, LAOF's optical-flow target~\cite{bu2026laof}, and CoMo's regularization~\cite{yang2026learning} into LAOM, following their released implementations.
All methods are evaluated using the same action-recovery and control protocols.

\paragraph{Real robot videos.}
We adapt UniVLA~\cite{bu2025univla} on six datasets: Bridge V2~\cite{walke2023bridgedata}, Berkeley Autolab UR5, Jaco Play, Taco Play~\cite{o2024open}, RT-1~\cite{brohan2022rt}, and IRASim~\cite{zhu2024irasim} (Section~\ref{cs:video}).
We compare against fine-tuned UniVLA and its adaptations with the same auxiliary guidance objectives, as well as MaskLAM's mask input and reconstruction loss.
We also evaluate released LAMs without fine-tuning~\cite{gao2026dreamdojo,wei2026causally,chen2026villa,chen2025moto}.

\paragraph{Implementation.}
The compact state is attention-pooled from an $8\times8$ feature grid on DCS and from $256$ patch tokens on real videos.
In both settings, the forward model predicts three future compact states from the same inferred latent action, and $\mathcal{L}_{\mathrm{dyn}}$ is averaged over the three targets.
These targets are at $t{+}1$, $t{+}2$, and $t{+}3$ on DCS and $t{+}k$, $t{+}2k$, and $t{+}3k$ on real videos, where $k$ is the interval between the two observations.
SMTD uses frozen DINOv2 patch features: \mbox{DINOv2-L} on DCS and UniVLA's \mbox{DINOv2-B} on real videos.
To balance gradient scales, we rescale the SMTD loss so that its backbone gradient norm is approximately $\lambda$ times that of the other loss terms.
We use $\lambda=0.1$ across all four DCS tasks and $\lambda=0.3$ across all six real robot datasets.
Within each setting, all methods we train share the same training budget: ten epochs on DCS and $10{,}000$ steps per dataset on real videos.
Sections~\ref{cs:sec} and~\ref{cs:sec:base} provide the remaining implementation details.

\subsection{Action recovery and context dependence}
\label{sec:exp:dcs}

\paragraph{Action recovery and control.}
We evaluate action recovery with linear probes and downstream control through returns of policies cloned on latent actions, following LAOM (Table~\ref{tab:dcs}).
Among methods trained without action labels, Agent-State LAM achieves the lowest action error on every task and the highest return in all eight task--stage pairs.
Across three independent training seeds (Section~\ref{cs:seeds}), Agent-State LAM reduces LAOM's mean action error by $57$--$61\%$ on hopper, cheetah, and walker and by $23\%$ on humanoid.
The mean $\pm$ standard deviation of action error is \mbox{$1.12\pm0.02$} vs.\ \mbox{$2.60\pm0.13$} on hopper, \mbox{$1.06\pm0.12$} vs.\ \mbox{$2.67\pm0.17$} on cheetah, \mbox{$2.37\pm0.10$} vs.\ \mbox{$6.04\pm1.03$} on walker, and \mbox{$0.1123\pm0.0010$} vs.\ \mbox{$0.1467\pm0.0011$} on humanoid, respectively.
With $32$k or $128$k action labels, LAOM improves linear-probe accuracy, while Agent-State LAM remains competitive in downstream control.

\begin{table*}[t]
\centering
\setlength{\tabcolsep}{6pt}%
\caption{\textbf{Action recovery on real robot videos.} Normalized test error ($\downarrow$) of linear probes predicting end-effector displacement from quantized ($z_q$) or continuous ($z$) latent actions. n/a indicates methods without a quantized latent.}
\label{tab:video}
\vspace{3pt}
\begin{tabular}{l cc cc cc cc cc cc}
\toprule
 & \multicolumn{2}{c}{Bridge} & \multicolumn{2}{c}{UR5} & \multicolumn{2}{c}{Jaco} & \multicolumn{2}{c}{Taco} & \multicolumn{2}{c}{RT-1} & \multicolumn{2}{c}{IRASim} \\
\cmidrule(lr){2-3} \cmidrule(lr){4-5} \cmidrule(lr){6-7} \cmidrule(lr){8-9} \cmidrule(lr){10-11} \cmidrule(lr){12-13}
 & $z_q$ & $z$ & $z_q$ & $z$ & $z_q$ & $z$ & $z_q$ & $z$ & $z_q$ & $z$ & $z_q$ & $z$ \\
\midrule
UniVLA, released & 0.78 & 0.52 & 0.46 & 0.21 & 0.56 & 0.39 & 0.83 & 0.65 & 0.77 & 0.53 & 0.80 & 0.59 \\
DreamDojo~\cite{gao2026dreamdojo}, released & n/a & 0.63 & n/a & 0.17 & n/a & 0.41 & n/a & 0.70 & n/a & 0.66 & n/a & 0.51 \\ %
CD-LAM~\cite{wei2026causally}, released & n/a & 0.62 & n/a & 0.17 & n/a & 0.40 & n/a & 0.69 & n/a & 0.66 & n/a & 0.51 \\ %
villa-X~\cite{chen2026villa}, released & 0.67 & 0.62 & 0.55 & 0.39 & 0.63 & 0.57 & 0.77 & 0.70 & 0.78 & 0.73 & 0.57 & 0.53 \\
Moto~\cite{chen2025moto}, released & 0.93 & 0.91 & 0.46 & 0.36 & 0.94 & 0.89 & 0.88 & 0.82 & 0.90 & 0.85 & 0.99 & 0.98 \\ %
\midrule
UniVLA, fine-tuned & 0.75 & 0.59 & 0.42 & 0.25 & 0.57 & 0.45 & 0.84 & 0.73 & 0.71 & 0.57 & 0.67 & 0.58 \\
\quad + Seq-$\Delta$REPA~\cite{jiang2026olaf} & 0.67 & 0.35 & 0.31 & \textbf{0.10} & 0.58 & \textbf{0.26} & 0.83 & 0.55 & 0.74 & 0.44 & 0.66 & 0.42 \\ %
\quad + optical-flow target~\cite{bu2026laof} & 0.71 & 0.42 & 0.34 & 0.15 & 0.50 & 0.33 & 0.77 & 0.61 & 0.64 & 0.47 & 0.75 & 0.48 \\  %
\quad + CoMo~\cite{yang2026learning} & 0.71 & 0.49 & 0.52 & 0.24 & 0.56 & 0.38 & 0.83 & 0.69 & 0.73 & 0.56 & 0.61 & 0.49 \\  %
\quad + MaskLAM~\cite{fechner2026segment} & 0.76 & 0.54 & 0.54 & 0.28 & 0.56 & 0.39 & 0.86 & 0.70 & 0.71 & 0.55 & 0.67 & 0.52 \\ %
\quad \textbf{+ Agent-State LAM} & \textbf{0.50} & \textbf{0.33} & \textbf{0.28} & 0.12 & \textbf{0.40} & \textbf{0.26} & \textbf{0.69} & \textbf{0.43} & \textbf{0.57} & \textbf{0.40} & \textbf{0.41} & \textbf{0.30} \\
\quad \textit{+ true actions} & \textit{0.28} & \textit{0.17} & \textit{0.14} & \textit{0.06} & \textit{0.17} & \textit{0.15} & \textit{0.49} & \textit{0.43} & \textit{0.40} & \textit{0.31} & \textit{0.35} & \textit{0.26} \\ %
\bottomrule
\end{tabular}
\end{table*}

\paragraph{Dependence on visual context.}
We test how the factors~\cone--\cfour{} affect latent actions and representations (Table~\ref{tab:dcs-change}; more metrics in Section~\ref{cs:sec:rep}).
Averaged over the four DCS tasks, Agent-State LAM achieves the highest pose R@1 (\mbox{Figure~\ref{fig:dcs-change}}), the lowest background-motion response share, and the smallest changes under static swaps and recoloring.

Agent masks lower both background measures, but MaskLAM needs them at test time, and selecting the agent does not separate its pose from its appearance~\cfour{}: its representation still changes when only the agent is recolored.
Adding LAOF's optical-flow target or CoMo's regularization to LAOM leaves background-motion response shares comparable to LAOM's.
Seq-$\Delta$REPA aligns latent actions with changes in frozen V-JEPA~2 features~\cite{assran2025v}; on DCS, where the distractor video fills most of the frame, these changes carry no linear information about the action (Section~\ref{cs:base}), and adding it to LAOM lowers pose R@1 from $40\%$ to $37\%$.
With $32$k or $128$k action labels, LAOM achieves $55$--$56\%$ R@1; Agent-State LAM achieves $63\%$ and lower context measures~\ctwo--\cfour{} without action labels during LAM training.

\subsection{Ablation and robustness studies}
\label{sec:exp:ablation}

\paragraph{Component analysis.}
We ablate the compact state and SMTD (Table~\ref{tab:ablation}; more variants in Section~\ref{cs:sec:abl}).
Without SMTD, compact-state prediction yields an action error of $3.32$ and pose R@1 of $10\%$, alongside strong background dependence.
Agent masks on the input partly remedy this failure (R@1 $40\%$, against $67\%$ for Agent-State LAM): they restrict the visible region, yet persistent appearance remains available within it.
To assess the role of compactness, we replace the inverse model's compact inputs with dense features: the action error rises from $1.14$ to $1.74$.
Replacing all compact states with dense features further increases the action error to $3.46$ and reduces pose R@1 from $62\%$ to $31\%$.
Together, these results highlight the importance of compact states for both action inference and prediction.

\begin{figure}[t]
\centering
\includegraphics[width=\columnwidth]{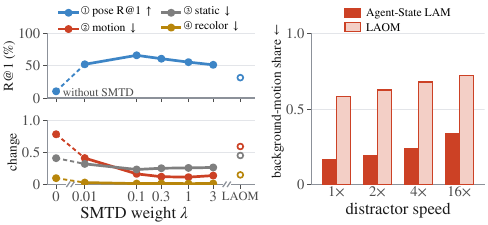}
\caption{\textbf{Sensitivity to SMTD weight and distractor speed} (DCS hopper). Left: pose retrieval~\cone{} and visual-context dependence~\ctwo--\cfour{} across SMTD weights~$\lambda$, with LAOM for comparison. Right: background-motion share~\ctwo{} as background videos are accelerated during training and evaluation; Agent-State LAM's share remains less than half of LAOM's at every tested speed.}
\label{fig:rho}
\end{figure}

\begin{figure*}[t]
\centering
\begin{subfigure}[b]{0.63\textwidth}
\centering
\includegraphics[width=\linewidth]{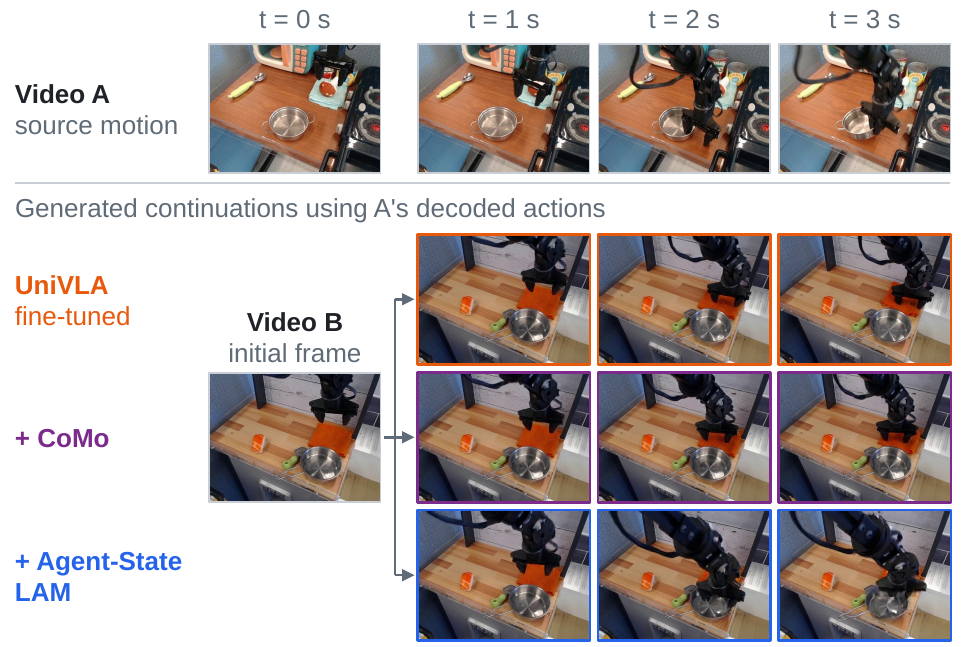}
\caption{Motion transfer across visual contexts.}
\label{fig:wm:transfer}
\end{subfigure}\hfill
\begin{subfigure}[b]{0.35\textwidth}
\centering
\includegraphics[width=\linewidth]{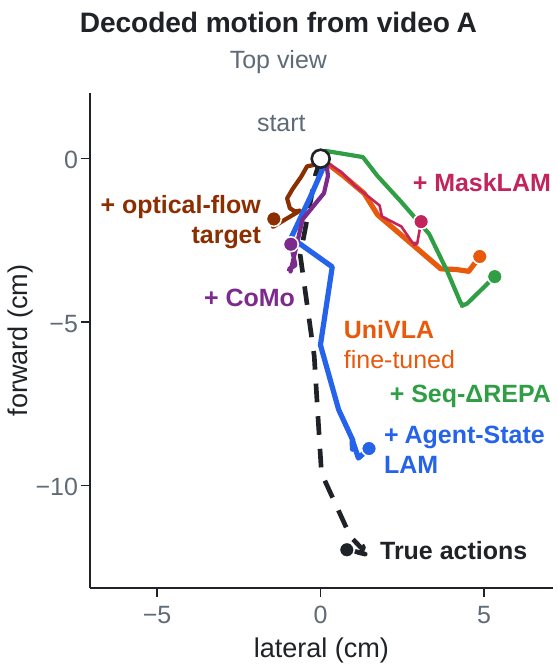}
\caption{Decoded end-effector motion.}
\label{fig:wm:path}
\end{subfigure}
\caption{\textbf{Latent actions drive a frozen world model.} (a) Cross-context motion transfer: latent actions inferred from IRASim video A are mapped to robot commands by a linear decoder fitted on training data. These commands drive frozen Cosmos-Predict2, initialized with the first frame of video B. With Agent-State LAM, the generated arm moves onto the pot, as in A; with fine-tuned UniVLA and its CoMo variant, the arm stays over the towel. (b) Top-view end-effector paths computed from actions decoded from video A, with true actions shown for reference. These paths are not tracked in the generated videos.}
\label{fig:wm}
\end{figure*}

\begin{table}[t]
\centering
\caption{\textbf{World-model prediction from decoded latent actions.} A linear decoder fitted on training data maps each model's latent actions to robot commands, which condition a frozen world model initialized from the real first frame. PSNR: comparison between the generated and ground-truth continuations; Traj.: \mbox{WorldArena} arm-trajectory accuracy~\cite{shang2026worldarena} measured on generated videos.}
\label{tab:wm}
\vspace{3pt}
\setlength{\tabcolsep}{1.5pt}
\begin{tabular}{@{}l cc cc cc@{}}
\toprule
 & \multicolumn{2}{c}{IRASim} & \multicolumn{2}{c}{Bridge} & \multicolumn{2}{c}{RT-1} \\
\cmidrule(lr){2-3} \cmidrule(lr){4-5} \cmidrule(l){6-7}
conditioning & PSNR & Traj. & PSNR & Traj. & PSNR & Traj. \\[-3pt]
 & $\uparrow$ & $\uparrow$ & $\uparrow$ & $\uparrow$ & $\uparrow$ & $\uparrow$ \\
\midrule
\textit{true actions} & \textit{26.05} & \textit{97.1} & \textit{19.73} & \textit{65.6} & \textit{23.73} & \textit{86.9} \\
\textit{other clip's actions} & \textit{21.46} & \textit{31.0} & \textit{17.05} & \textit{22.0} & \textit{19.51} & \textit{21.6} \\
\midrule
UniVLA, released & 21.96 & 41.9 & 18.15 & 37.8 & 21.10 & 47.9 \\
UniVLA, fine-tuned & 22.38 & 51.1 & 18.24 & 41.6 & 21.18 & 48.2 \\
+ Seq-$\Delta$REPA & 22.43 & 51.2 & 18.44 & 44.9 & 20.99 & 47.1 \\
+ optical-flow target & 21.89 & 39.5 & 18.50 & 42.6 & 21.09 & \textbf{56.4} \\
+ CoMo & 22.63 & 56.8 & 18.49 & 43.0 & 21.20 & 51.2 \\
+ MaskLAM & 22.45 & 53.2 & 18.17 & 39.7 & 21.12 & 48.6 \\
\textbf{+ Agent-State LAM} & \textbf{23.35} & \textbf{67.5} & \textbf{18.63} & \textbf{45.8} & \textbf{21.37} & 53.3 \\
\bottomrule
\end{tabular}
\end{table}

\paragraph{SMTD analysis.}
Without temporal contrast (positives only), SMTD provides no incentive to reject shared context and performs worse than omitting it.
With temporal contrast retained, frame-mean scoring degrades action recovery and pose retrieval, supporting the importance of spatially flexible local matching.
State-to-state contrast also underperforms full SMTD, whereas using the smaller DINOv2-S teacher retains most of the gains.

\paragraph{SMTD weight and distractor speed.}
We further examine whether these benefits persist across different regularization strengths and background dynamics.
All tested SMTD weights from $0.1$ to $3$ improve pose retrieval~\cone{} and lower the context measures~\ctwo--\cfour{} relative to LAOM (Figure~\ref{fig:rho}, left).
To test robustness to faster distractors, we accelerate background videos during both training and evaluation.
At every tested speed up to $16\times$, our background-motion response share remains below half of LAOM's (Figure~\ref{fig:rho}, right).
The action-recovery advantage also persists, with errors of $1.50$ versus LAOM's $3.40$ at $16\times$.

\subsection{Action recovery from real robot videos}
\label{sec:exp:video}

We test whether agent-state learning extends beyond DCS to a different LAM architecture and six real robot datasets.
Linear probes predict end-effector displacement from the quantized latent $z_q$, used by UniVLA's policy, and the continuous latent $z$ on test episodes (Table~\ref{tab:video}).

Although MaskLAM improves action recovery over LAOM on most DCS tasks, its UniVLA adaptation does not improve quantized-latent error on five of the six datasets.
In contrast, Agent-State LAM reduces fine-tuned UniVLA's error for both latents on every dataset, extending its benefits to real robot videos and a different feature-space LAM.

\subsection{World-model prediction with latent actions}
\label{sec:exp:wm}

To assess the utility of learned latent actions for world-model prediction, we fit a linear decoder on training data to map each model's latents to robot commands.
Starting from each held-out clip's first frame, these commands condition frozen world models: Cosmos-Predict2~\cite{ali2025world} on IRASim and Cosmos~3~\cite{agarwal2026cosmos} on Bridge and RT-1.
Because whole-frame similarity can obscure motion errors, we complement PSNR with \mbox{WorldArena}'s arm-trajectory accuracy~\cite{shang2026worldarena}.

Agent-State LAM achieves the highest PSNR among compared LAMs on all three datasets and the highest mean trajectory accuracy on IRASim and Bridge; the optical-flow-target variant scores higher on RT-1 (Table~\ref{tab:wm}; command recovery and confidence intervals in Section~\ref{cs:sec:wm}).
Figure~\ref{fig:wm} further illustrates cross-context motion transfer.

\section{Conclusion and discussion}
\label{sec:conclusion}

We introduced Agent-State LAM to reduce visual-context entanglement by jointly learning latent actions and a compact state shared by action inference and forward prediction.
SMTD guides what this state retains through spatially flexible matching and nearby-frame discrimination, favoring the agent's evolving physical state.
Experiments on DCS and real robot videos demonstrate improved action recovery, downstream control, and world-model prediction.

\paragraph{Limitations and future work.}
Despite substantial improvements in end-effector translation recovery, recovering fine-grained actions such as wrist rotation and gripper commands remains challenging.
Although Agent-State LAM achieves higher rotation-recovery scores than fine-tuned UniVLA across the three world-model datasets, with the largest gain on \mbox{RT-1}, absolute performance remains limited (Section~\ref{cs:wm}).
These action components may induce subtle or ambiguous visual changes that are hard to infer from video alone.
Future work may explore complementary supervision, such as language guidance, to improve recovery of these actions.

\FloatBarrier
{
    \small
    \bibliographystyle{ieeenat_fullname}
    \bibliography{main}
}

\clearpage
\makeatletter\suppfloatpages\suppcontents\makeatother
\maketitlesupplementary
\setcounter{section}{0}\setcounter{table}{0}\setcounter{figure}{0}\setcounter{equation}{0}
\renewcommand{\theHsection}{S\arabic{section}}
\renewcommand{\theHtable}{S\arabic{table}}
\renewcommand{\theHfigure}{S\arabic{figure}}
\renewcommand{\theHequation}{S\arabic{equation}}
\renewcommand{\thesection}{S\arabic{section}}
\renewcommand{\thetable}{S\arabic{table}}
\renewcommand{\thefigure}{S\arabic{figure}}
\renewcommand{\theequation}{S\arabic{equation}}
\noindent\textbf{Overview.} For clarity and conciseness, we organize this supplement into five thematic sections (\ref{cs:sec}--\ref{cs:sec:wm}).
Section~\ref{cs:sec} describes the implementation of Agent-State LAM on the Distracting Control Suite (DCS) and real robot video, and Section~\ref{cs:sec:base} presents the baseline implementations, their validation and diagnostics, and comparisons under matched settings.
Section~\ref{cs:sec:rep} details the controlled-intervention metrics, tests robustness under natural re-renderings, analyzes spatial matching in SMTD, and evaluates arm-state retrieval on real robot video.
Section~\ref{cs:sec:abl} reports further ablations on Bridge and DCS and results across training seeds, and Section~\ref{cs:sec:wm} evaluates world-model prediction, motion transfer, and command recovery, with qualitative examples.

{%
\setcounter{tocdepth}{2}
\makeatletter
\renewcommand\numberline[1]{#1\hskip 1em}
\medskip\noindent{\large\sffamily\bfseries Contents}\par\vspace{2pt}
\renewcommand*\l@section[2]{%
  \addpenalty\@secpenalty
  \addvspace{1.0em \@plus\p@}%
  \setlength\@tempdima{1.5em}%
  \begingroup
    \parindent \z@ \rightskip \@pnumwidth \@plus 1fil \hyphenpenalty\@M
    \parfillskip -\@pnumwidth
    \leavevmode \bfseries
    \advance\leftskip\@tempdima
    \hskip -\leftskip
    #1\nobreak\hfill \nobreak\hb@xt@\@pnumwidth{\hss #2 \kern-\p@\kern\p@}\par
  \endgroup}
\@starttoc{apptoc}
\makeatother
}
\begin{table*}[t]
\centering
\setlength{\tabcolsep}{6pt}
\caption{\textbf{Settings of Agent-State LAM.} Top: the base model and training setup, which follow LAOM~\cite{nikulin2025latent} on DCS and the fine-tuned UniVLA~\cite{bu2025univla} on real robot video, with training budgets shared across the models we train. Bottom: the compact state and SMTD. $k$ is the frame interval on real video (Table~\ref{cs:tab:datasets}); $w_{\mathrm{dyn}}$, $\lambda$, $\tau_p$, and $\tau$ are the weights and temperatures of Eqs.~\eqref{eq:method:score}--\eqref{eq:method:objective}.}
\label{cs:tab:settings}
\vspace{3pt}
\begin{tabular}{@{}l>{\raggedright\arraybackslash}p{0.35\textwidth}>{\raggedright\arraybackslash}p{0.36\textwidth}@{}}
\toprule
 & DCS & Real robot video \\
\midrule
\multicolumn{3}{@{}l}{\textit{Base model and training setup}} \\
\hspace{0.6em}input & three-frame stacks at $64\times64$; pairs $1$--$10$ steps apart & $256$ frozen DINOv2-B patch tokens per frame; pairs $k$ frames apart \\
\hspace{0.6em}dense features & $N=64$, $c=192$ ($8\times8$ grid; $Nc=12{,}288$) & $N=256$, $c=768$ ($Nc=196{,}608$) \\
\hspace{0.6em}latent action & continuous, $8{,}192$-dimensional & four tokens of sixteen codes each \\
\hspace{0.6em}batch size & $512$ & $512$ \\
\hspace{0.6em}learning rate & $10^{-4}$; $3$-epoch warm-up, then linear decay & $10^{-5}$ (pretrained) / $10^{-4}$ (new or adapted); $500$-step warm-up, then linear decay \\
\hspace{0.6em}training budget & ten epochs ($96{,}680$ updates) & $10{,}000$ steps \\
\midrule
\multicolumn{3}{@{}l}{\textit{Compact state and SMTD}} \\
\hspace{0.6em}compact state & $d=384$ & $d=768$ \\
\hspace{0.6em}prediction targets & EMA states at $t{+}1$, $t{+}2$, $t{+}3$ ($w_{\mathrm{dyn}}=1$) & EMA states at $t{+}k$, $t{+}2k$, $t{+}3k$ ($w_{\mathrm{dyn}}=3$; unit weight per offset) \\
\hspace{0.6em}target encoder & EMA, decay $0.999$ per update & EMA, decay $0.999$ per update \\
\hspace{0.6em}SMTD reference encoder & frozen DINOv2-L patch features & frozen DINOv2-B patch tokens \\
\hspace{0.6em}SMTD weight, temperatures & $\lambda=0.1$, $\tau_p=0.05$, $\tau=0.07$ & $\lambda=0.3$, $\tau_p=0.05$, $\tau=0.07$ \\
\bottomrule
\end{tabular}
\end{table*}
\section{Implementation of Agent-State LAM}
\label{cs:sec}
Table~\ref{cs:tab:settings} summarizes the implementation settings of Agent-State LAM on DCS and real robot video.
Across both benchmarks, we retain the architecture, input format, latent parameterization, and training budget of the base model (LAOM~\cite{nikulin2025latent} on DCS and the fine-tuned UniVLA~\cite{bu2025univla} on real robot video), adding attention pooling to form the shared compact state and training jointly with three-step EMA state prediction and SMTD.
The SMTD temperatures, the EMA decay, and the three-step horizon are the same on both benchmarks (Table~\ref{cs:tab:settings}), and a single $\lambda$ per benchmark balances $\mathcal{L}_{\mathrm{SMTD}}$'s backbone gradient norm relative to the other loss terms (Section~\ref{sec:exp:setup}).
Full training configurations are released with \mbox{the code.}

\subsection{DCS}
\label{cs:dcs}
We use the training and test files released with LAOM~\cite{nikulin2025latent} for hopper-hop, cheetah-run, walker-run, and humanoid-walk, with $5{,}000$ training trajectories of $1{,}000$ steps per task and natural DAVIS~\cite{Pont-Tuset_arXiv_2017} backgrounds under the camera and color perturbations of DCS~\cite{stone2021distracting}.
We build on LAOM's released codebase, retaining its backbone, dynamics networks, and optimization setup; all models are trained for ten epochs and evaluated at \mbox{the final checkpoint.}

To form the compact agent state, each spatial backbone feature is projected to the compact-state dimension with a learned positional embedding and aggregated by a single cross-attention layer with one learned query followed by a residual MLP; the EMA target encoder tracks both the backbone and attention pooling.
The inverse model operates on the compact states of frame pairs $(t, t{+}k)$, with $k\in[1,10]$ as in LAOM, while the forward model, with LAOM's architecture, predicts the EMA compact states at $t{+}1$, $t{+}2$, and $t{+}3$ (computed from unaugmented frames) for every $k$.
For SMTD, frozen DINOv2-L~\cite{oquab2023dinov2} patch features are projected by a fixed PCA, fitted on unlabeled training frames, to the $384$-dimensional width of the compact state, so that $q$ is the identity (the $384$-dimensional DINOv2-S teacher of Table~\ref{tab:ablation} needs no PCA), and the negative-sampling window narrows from $8$--$25$ to $1$--$4$ steps over training, from \mbox{easier to harder negatives.}

For Stage-1 action recovery (Table~\ref{tab:dcs}), every model is evaluated with the same ridge regression probe ($\alpha=1$, standardized features) mapping latent actions to LAOM's action targets (the expert's raw outputs, before they are clipped to the action bounds), fitted on $4{,}096$ labeled training transitions ($32$ fixed time steps from each of $128$ trajectories) and scored by test MSE on the released test files.
Downstream control (Stages~2--3) follows LAOM's policy-cloning protocol and default evaluation counts identically for every method: a policy is cloned on each model's latent actions without action labels over a shared training collection generated with the same DCS simulator settings, and its outputs are decoded by a decoder fitted on all action labels (Stage~2, $5$ evaluation episodes) or on $32$ labeled trajectories (Stage~3, $25$ evaluation episodes), with returns normalized by LAOM's behavior-cloning reference values (multi-seed results in Section~\ref{cs:seeds}).

\subsection{Real robot video}
\label{cs:video}
Table~\ref{cs:tab:datasets} lists the six datasets: Bridge~V2~\cite{walke2023bridgedata}, UR5, Jaco Play, and Taco Play~\cite{o2024open}, RT-1~\cite{brohan2022rt}, and IRASim~\cite{zhu2024irasim}.
Test episodes are held out prior to adaptation and excluded from training: a random fifth of the source episode chunks on Bridge and RT-1 ($6$ of $33$ and $17$ of $88$), a random fifth of the episodes on UR5, Jaco, and Taco, and IRASim's released test split (clips of at least $12$ frames, $26\%$ of clips).
Because the released UniVLA checkpoint was pretrained on Bridge, RT-1, Taco, Jaco, and UR5, and the IRASim clips are Bridge recordings, test episodes of all six datasets may have appeared in its pretraining; this applies identically to every UniVLA-based row.
\suppressfloats[t]
\begin{table}[t]
\centering
\setlength{\tabcolsep}{5pt}
\caption{\textbf{Real-video datasets.} Total episodes (clips on IRASim), held-out test episodes, frame interval $k$ between paired observations, and labeled pairs used by the action-supervised reference. Recording rates vary across datasets ($3$--$15$~Hz), and consecutive frames on the higher-rate datasets (Jaco, Taco) show very small arm displacement, so $k$ differs by dataset; with these values, paired observations are a comparable $0.6$--$1.0$~s apart, except on IRASim, where $k=1$ matches single-step world-model conditioning. Every model in Table~\ref{tab:video} infers its latent from the same $k$-frame pairs.}
\label{cs:tab:datasets}
\vspace{3pt}
\begin{tabular}{@{}lcccc@{}}
\toprule
 & episodes & test & $k$ & labeled pairs \\
\midrule
Bridge V2 & $5{,}954$ & $1{,}138$ & $3$ & $157$k \\
UR5 & $896$ & $179$ & $5$ & $67$k \\
Jaco Play & $976$ & $195$ & $9$ & $49$k \\
Taco Play & $3{,}242$ & $648$ & $9$ & $148$k \\
RT-1 & $6{,}122$ & $1{,}185$ & $3$ & $203$k \\
IRASim & $6{,}736$ & $1{,}737$ & $1$ & $165$k \\
\bottomrule
\end{tabular}
\end{table}

UniVLA's latent action model~\cite{bu2025univla} encodes frozen DINOv2-B patch tokens of two frames with a $12$-block spatio-temporal transformer and quantizes the latent action into discrete codes.
We split this pretrained encoder at block~$4$: blocks~$1$--$4$ followed by attention pooling produce the per-frame compact state, blocks~$5$--$12$ infer the latent action from pairs $k$ frames apart, and UniVLA's decoder predicts the EMA compact states at $t{+}k$, $t{+}2k$, and $t{+}3k$, with UniVLA's codebook loss retained.
For SMTD, a projection MLP $q$ maps the compact state into the frozen DINOv2-B token space, and negatives are sampled from nearby frames within each episode.
Following standard fine-tuning practice (Table~\ref{cs:tab:settings}, shared with the fine-tuned baselines in Section~\ref{cs:base}), pretrained weights that keep their original role use a learning rate of $10^{-5}$, and the action-code heads and new or adapted layers use $10^{-4}$; unlike on DCS, where the model is trained from scratch, the attention pooling here is newly initialized inside a pretrained transformer, so the new and adapted layers are first warm-started from the released UniVLA latent for $1{,}000$ of the \mbox{$10{,}000$ steps.}

For evaluation (Table~\ref{tab:video}), every model infers its latent from the same two observations, $k$ frames apart.
Ridge probes on the continuous and quantized latents ($z$ and $z_q$) are fitted on up to $20{,}000$ training pairs---with regularization selected on held-out training episodes---and evaluated on up to $15{,}000$ test pairs.
The target is end-effector translation summed over the $k$-frame gap (from commanded actions, or recorded end-effector states on IRASim), scored by MSE normalized by the test-target variance.

\section{Baseline implementations and comparisons}
\label{cs:sec:base}
This section describes the implementation of the baselines used in our experiments and presents additional comparisons and diagnostics of their performance.
\subsection{Baseline implementations}
\label{cs:base}
The baselines were proposed with different architectures, data, and training code, and one of them has no public code.
To compare them under one protocol, we run the released code where it applies to our setting and otherwise re-implement the method or add its objective to a common model.
The LAPO~\cite{schmidt2024learning} and LAOM~\cite{nikulin2025latent} rows of Table~\ref{tab:dcs} run LAOM's released code with its configurations, and the label-supervised references add $32{,}000$ or $128{,}000$ labeled transitions ($0.6\%$ and $2.6\%$ of the training transitions) to LAOM as an auxiliary target.
On real video, the released models of Table~\ref{tab:video} (UniVLA~\cite{bu2025univla}, DreamDojo~\cite{gao2026dreamdojo}, CD-LAM~\cite{wei2026causally}, villa-X~\cite{chen2026villa}, and Moto~\cite{chen2025moto}) are evaluated as released, and every fine-tuned row starts from the released UniVLA and is trained for the same $10{,}000$ steps with the learning-rate rule \mbox{of Section~\ref{cs:video}.}

The MaskLAM paper~\cite{fechner2026segment} links no code, so we implemented the method from its description: every input frame receives its agent mask as an additional channel, and the reconstruction loss is restricted to the agent's pixels in the target frame.
We train it in LAOM's setup (optimizer, augmentations, latent size, and hidden width of the heads) with MaskLAM's cosine schedule.
For policy cloning and for the measurements that need exact agent masks, we generated a second collection with the simulator of LAOM's released DCS files, which also stores the simulator's agent masks (the mask-bearing collection).
Every model's policy is cloned on the training set of this collection: MaskLAM needs a mask for each frame that it labels, and the other models, which read no masks, are cloned on the same frames.
With exact masks on hopper, trained and scored on the mask-bearing collection, the LAOM-based setup reaches an action error of $2.06$ and the hyperparameters of the MaskLAM paper $3.55$ at best, so Table~\ref{tab:dcs} uses the stronger setup.
The released DCS files carry no masks, so the masks come from SAM~2.1~\cite{ravi2025sam}, the model of the MaskLAM paper, prompted with one fixed box and point per task on the last frame of each trajectory and tracked backward across each trajectory, with automatic re-prompting of implausible tracks.
Against the simulator's masks on the mask-bearing collection, these masks reach a mean IoU of $0.79$ / $0.79$ / $0.81$ / $0.85$ on hopper / cheetah / walker / humanoid.

On real video, arm masks are extracted with Grounding~DINO~\cite{liu2024grounding} and SAM~2.1 and tracked through each episode; whenever an initial detection misses the moving arm, we select the highest-motion candidate track (tracking the visible gripper on RT-1).
The fraction of arm pixels per token scales a learned embedding that is added to the input tokens and weights the reconstruction loss of the target frame's tokens, and the masks are also used \mbox{at test time.}

Seq-$\Delta$REPA~\cite{jiang2026olaf}, LAOF~\cite{bu2026laof}, and CoMo~\cite{yang2026learning} release code, each for its own latent action model and data, so the released implementations differ in more than the objective.
LAOF's, for instance, builds on LAPO, with a quantized latent and a U-Net flow decoder, and reaches an action error of $14.15$ on hopper, near LAPO's $16.55$ (against $2.35$ when its objective is ported onto LAOM).
We therefore add each method's objective to one common model per setting, LAOM on DCS and the fine-tuned UniVLA on real video, with the settings of its source except where stated.

The Seq-$\Delta$REPA rows align a linear head on the latent, by cosine similarity, with the normalized difference of the two frames' pooled V-JEPA~2 ViT-g features~\cite{assran2025v}, with the authors' weight of $0.02$ on DCS and a weight of $1$ on real video, which gives this baseline a lower error on Bridge than the authors' weight.
On DCS, where the distractor video fills most of the frame, this target carries no linear information about the action: a ridge probe from the target to the true action reaches $R^2\le0.00$ on all four tasks, at gaps of one and five steps ($4{,}800$ training transitions per task, five folds by trajectory).
On real video, the same probe reaches $0.40$ on Bridge, $0.71$ on UR5, and $0.24$ on IRASim for the end-effector translation, and there the term lowers the error of the fine-tuned UniVLA's continuous latent on every dataset \mbox{(Table~\ref{tab:video}).}

The optical-flow rows follow LAOF: RAFT~\cite{teed2020raft} flow is drawn as an image with LAOF's mapping and, on DCS, multiplied by the agent mask of the MaskLAM row, as LAOF does under moving distractors; a second head with the forward model's architecture predicts, from the latent, the features of these flow images (the EMA encoder's on DCS, frozen DINOv2-B features on real video).
On DCS, this term is weighted as SMTD is (Section~\ref{cs:sec}), which gives this baseline a lower error on every task than LAOF's weight of $1$; on real video it keeps LAOF's weight.
The CoMo rows give the inverse model CoMo's temporal-difference input and add its contrastive term with time-reversed negatives, with the temperature $0.1$ and the weight $0.004$ of its released code.
Finally, the action-supervised reference of Table~\ref{tab:video} (+ true actions) adds a regression from the quantized latent to the true action \mbox{of every training pair.}

\suppressfloats[t]
\subsection{Baseline validation and diagnostics}
\label{cs:valid}
\begin{table}[t]
\centering
\setlength{\tabcolsep}{3.5pt}
\caption{\textbf{MaskLAM validation under its paper's probe.} Normalized linear-probe error $\downarrow$ on clipped actions, as in Table~12 of~\cite{fechner2026segment}; our probes are fitted on $50{,}000$ labels. Our Table~\ref{tab:dcs} models are trained and probed on LAOM's released files, and~\cite{fechner2026segment} releases neither code nor checkpoints, so the top block gives the published values, measured on that paper's own data, as a reference. The bottom block scores our Table~\ref{tab:dcs} models on LAOM's test files, and our exact-mask MaskLAM on held-out trajectories of our mask-bearing collection (Section~\ref{cs:base}).}
\label{cs:tab:published}
\vspace{3pt}
\begin{tabular*}{\columnwidth}{@{\extracolsep{\fill}}l cccc@{}}
\toprule
 & hop. & cheet. & walk. & hum. \\
\midrule
\multicolumn{5}{@{}l}{\textit{Published (MaskLAM paper, Table~12; its own data)}} \\
\hspace{0.6em}MaskLAM, exact masks & 0.185 & 0.117 & 0.169 & 0.353 \\
\hspace{0.6em}MaskLAM, SAM masks & 0.260 & 0.115 & 0.170 & 0.385 \\
\midrule
\multicolumn{5}{@{}l}{\textit{Our runs, same probe}} \\
\hspace{0.6em}MaskLAM, exact masks & 0.148 & 0.114 & 0.121 & 0.388 \\
\hspace{0.6em}MaskLAM, SAM masks & 0.160 & 0.125 & 0.142 & 0.414 \\
\hspace{0.6em}Agent-State LAM & \textbf{0.082} & \textbf{0.078} & \textbf{0.100} & \textbf{0.194} \\
\bottomrule
\end{tabular*}
\end{table}
\begin{table}[t]
\centering
\setlength{\tabcolsep}{4pt}
\caption{\textbf{Forms of each baseline that we trained} (DCS hopper, action error $\downarrow$ with the probe of Table~\ref{tab:dcs}). Bold: the form reported in Table~\ref{tab:dcs}; the last three rows repeat Table~\ref{tab:dcs} for reference. $^\dagger$With exact masks, trained and scored on our mask-bearing collection; these values compare with \mbox{one another only.}}
\label{cs:tab:variants}
\vspace{3pt}
\begin{tabular*}{\columnwidth}{@{\extracolsep{\fill}}l c@{}}
\toprule
\multicolumn{2}{@{}l}{\textit{MaskLAM, exact masks$^\dagger$}} \\
\hspace{0.6em}paper's settings, mask in the loss & 3.55 \\
\hspace{0.6em}paper's settings, mask input and loss & 4.48 \\
\hspace{0.6em}LAOM's setup, heads without a hidden layer & 3.64 \\
\hspace{0.6em}LAOM's setup & 2.06 \\
\midrule
\multicolumn{2}{@{}l}{\textit{MaskLAM, SAM masks}} \\
\hspace{0.6em}first masks & 3.61 \\
\hspace{0.6em}re-prompted masks & \textbf{2.41} \\
\midrule
\multicolumn{2}{@{}l}{\textit{LAOM + Seq-$\Delta$REPA}} \\
\hspace{0.6em}authors' weight & \textbf{3.16} \\
\hspace{0.6em}weight $1$ & 7.17 \\
\hspace{0.6em}authors' weight, DINOv2-L teacher & 6.64 \\
\midrule
\multicolumn{2}{@{}l}{\textit{Optical-flow target}} \\
\hspace{0.6em}LAOF's released code, full-frame flow & 14.15 \\
\hspace{0.6em}LAOF's released code, masked flow$^\dagger$ & 12.71 \\
\hspace{0.6em}on LAOM, full-frame flow, LAOF's weight & 12.33 \\
\hspace{0.6em}on LAOM, masked flow, LAOF's weight & 2.46 \\
\hspace{0.6em}on LAOM, masked flow, weighted as SMTD & \textbf{2.35} \\
\midrule
\multicolumn{2}{@{}l}{\textit{LAOM + CoMo}} \\
\hspace{0.6em}temporal-difference input only & 2.35 \\
\hspace{0.6em}with the contrastive term & \textbf{2.12} \\
\midrule
LAPO & 16.55 \\
LAOM & 2.70 \\
Agent-State LAM & 1.14 \\
\bottomrule
\end{tabular*}
\end{table}
\begin{figure*}[t]
\centering
\includegraphics[width=\textwidth]{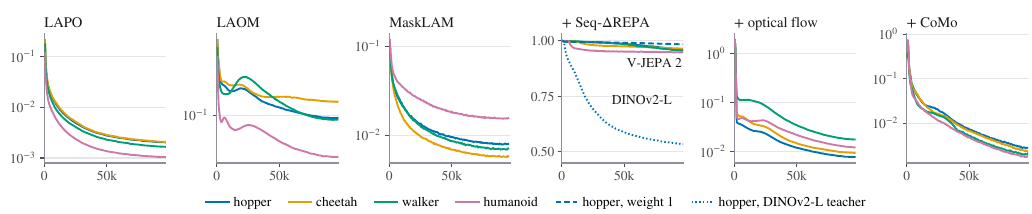}
\caption{\textbf{Training loss curves of the DCS baselines.} Loss over ten epochs (means over $500$ updates): pixel loss (LAPO, MaskLAM), feature-prediction loss (LAOM), and the terms added to LAOM by Seq-$\Delta$REPA ($1-\cos$ to its target), the optical-flow target, and CoMo (logarithmic vertical axes except Seq-$\Delta$REPA). All decrease substantially except Seq-$\Delta$REPA with V-JEPA~2 ($1.00 \to 0.95$--$0.96$; $0.98$ at weight~$1$ on hopper); the same alignment implementation with a DINOv2-L teacher falls to $0.54$ (while raising hopper error to $6.64$, Table~\ref{cs:tab:variants}), indicating that the term is optimized as intended and the flat curves reflect its V-JEPA~2 target on DCS (Sections~\ref{cs:base},~\ref{cs:valid}).}
\label{cs:fig:base-training}
\end{figure*}

\begin{table}[t]
\centering
\setlength{\tabcolsep}{3.5pt}
\caption{\textbf{Baseline comparisons under matched settings on DCS} (action error $\downarrow$ with the probe of Table~\ref{tab:dcs}). For each baseline, the first row is its Table~\ref{tab:dcs} result and the second row retrains it with Agent-State LAM's target and loss-balancing settings (Section~\ref{cs:matched}). Every baseline has the lower error in its original configuration on every task, which is the one used in Table~\ref{tab:dcs}. $^\dagger$For MaskLAM, the matched settings are LAOM's residual head designs and learning-rate schedule.}
\label{cs:tab:matched}
\vspace{3pt}
\begin{tabular*}{\columnwidth}{@{\extracolsep{\fill}}l cccc@{}}
\toprule
 & hop. & cheet. & walk. & hum. \\
\midrule
LAOM~\cite{nikulin2025latent} & 2.70 & 2.47 & 5.49 & 0.15 \\
\hspace{0.6em}matched settings & 4.30 & 3.48 & 10.63 & 0.23 \\
+ Seq-$\Delta$REPA~\cite{jiang2026olaf} & 3.16 & 2.45 & 7.10 & 0.14 \\
\hspace{0.6em}matched settings & 3.25 & 2.93 & 9.34 & 0.18 \\
+ optical-flow target~\cite{bu2026laof} & 2.35 & 2.32 & 3.06 & 0.14 \\
\hspace{0.6em}matched settings & 2.57 & 3.03 & 3.89 & 0.18 \\
+ CoMo~\cite{yang2026learning} & 2.12 & 2.07 & 4.18 & 0.14 \\
\hspace{0.6em}matched settings & 3.25 & 3.12 & 5.16 & 0.18 \\
MaskLAM~\cite{fechner2026segment} & 2.41 & 2.02 & 5.23 & 0.22 \\
\hspace{0.6em}matched settings$^\dagger$ & 4.19 & 3.13 & 8.75 & 0.35 \\
\midrule
Agent-State LAM & \textbf{1.14} & \textbf{1.01} & \textbf{2.37} & \textbf{0.11} \\
\bottomrule
\end{tabular*}
\end{table}
Because LAPO and LAOM use LAOM's official released code directly, our validation focuses on the baselines that we re-implemented or ported onto a shared model.
MaskLAM provides no public code or checkpoints, so Table~\ref{cs:tab:published} checks our implementation against its published numbers under the MaskLAM paper's own clipped-action probe.
With exact simulator masks, our implementation achieves lower error than the published values on hopper, cheetah, and walker ($0.148$, $0.114$, and $0.121$ vs.\ $0.185$, $0.117$, and $0.169$) and stays close on humanoid ($0.388$ vs.\ $0.353$).
With SAM masks, our Table~\ref{tab:dcs} model outperforms the published error on hopper and walker ($0.160$ and $0.142$ vs.\ $0.260$ and $0.170$) and remains within $0.03$ on cheetah and humanoid.
Although the two evaluations use different data collections, these comparisons indicate that our implementation matches the performance level of the published method.
Under MaskLAM's clipped-action probe as well, Agent-State LAM achieves the lowest error on every task, reducing MaskLAM's SAM-mask error by $29$--$53\%$, computed before rounding, across the four tasks \mbox{($0.078$--$0.194$ vs.\ $0.125$--$0.414$).}

Next, we verify that Table~\ref{tab:dcs} evaluates strong configurations of each baseline and examine how each baseline objective is optimized.
Across the DCS hopper variants compared in Table~\ref{cs:tab:variants}, Table~\ref{tab:dcs} uses the strongest configuration of each baseline: adopting LAOM's setup substantially improves MaskLAM over its paper's hyperparameters ($2.06$ vs.\ $3.55$ with exact masks) and re-prompted SAM masks improve over the first masks ($2.41$ vs.\ $3.61$), porting LAOF's masked optical-flow target onto LAOM with gradient-scale balancing improves sharply over LAOF's released code, which builds on LAPO ($2.35$ vs.\ $14.15$), CoMo's contrastive loss improves over its temporal-difference input alone ($2.12$ vs.\ $2.35$), and increasing the weight or switching to a DINOv2-L teacher degrades Seq-$\Delta$REPA ($7.17$ and $6.64$ vs.\ $3.16$).
Figure~\ref{cs:fig:base-training} confirms that every baseline objective decreases over the ten training epochs: LAPO's pixel loss and the optical-flow and CoMo terms drop by roughly two orders of magnitude, MaskLAM's masked pixel loss drops by $7$--$18\times$, and LAOM's prediction loss drops by $4$--$9\times$.
For Seq-$\Delta$REPA, the alignment loss decreases only modestly with the authors' V-JEPA~2 teacher, and increasing the weight to $1$ does not reduce it further while raising action error to $7.17$; switching to a DINOv2-L teacher yields much closer alignment yet raises action error to $6.64$, suggesting that tighter alignment to background-dominated feature differences degrades action recovery on DCS.

Finally, we check whether each ported objective improves the base model it is added to.
On DCS, both the optical-flow target and CoMo reduce LAOM's action error on all four tasks (Table~\ref{tab:dcs}); the exception is Seq-$\Delta$REPA, whose V-JEPA~2 target carries no linear action information under full-frame video distractors (Section~\ref{cs:base}).
On real robot video, Seq-$\Delta$REPA, the optical-flow target, and CoMo each lower fine-tuned UniVLA's continuous-latent error on all six datasets, while their quantized-latent error changes in both directions (Table~\ref{tab:video}).
Likewise, MaskLAM improves over unmasked pixel reconstruction (LAPO) on all four DCS tasks and lowers fine-tuned UniVLA's continuous-latent error on five of the six real-video datasets.
As a sanity check on the shared trainer used to integrate these auxiliary objectives, running plain LAOM through it achieves $2.53$ on hopper, matching LAOM's released code across three seeds ($2.60\pm0.13$).

\subsection{Comparisons under matched settings}
\label{cs:matched}
\begin{table*}[t]
\centering
\setlength{\tabcolsep}{2.2pt}%
\caption{\textbf{Natural re-rendering tests on DCS.} Change (\%) $\downarrow$ of each method's per-frame representation between two simulator renderings of the same held-out frame, relative to the distance between random frames of different trajectories: with another background video, and with a recolored agent while pose, camera, and background stay fixed. Last row, a separate measure without re-rendering: share of the background's frame-to-frame change pattern that a linear probe recovers from LAPO's latent action ($R^2$, \%, $\downarrow$). Underlined: quoted in the introduction. Bold: lowest; italic: label-supervised references. Right: cheetah and walker frames in the three renderings.}
\label{cs:tab:natural}
\vspace{3pt}
\begin{tabular}{@{}l ccccc@{\hspace{9pt}}ccccc@{}}
\toprule
 & \multicolumn{5}{c}{new background video $\downarrow$} & \multicolumn{5}{c}{recolored agent $\downarrow$} \\
\cmidrule(r){2-6}\cmidrule{7-11}
 & hop. & cheet. & walk. & hum. & mean & hop. & cheet. & walk. & hum. & mean \\
\midrule
LAPO~\cite{schmidt2024learning} & 91 & 95 & 63 & 75 & 81 & \textbf{1} & \textbf{1} & 3 & 2 & 2 \\
MaskLAM~\cite{fechner2026segment} & 19 & 23 & 5 & 5 & 13 & 7 & \underline{20} & 9 & 2 & \underline{9} \\
LAOM~\cite{nikulin2025latent} & 17 & 17 & \underline{38} & 9 & 20 & 15 & 14 & 16 & 14 & 15 \\
\quad + Seq-$\Delta$REPA~\cite{jiang2026olaf} & 32 & 20 & 43 & 9 & 26 & 22 & 17 & 32 & 15 & 22 \\
\quad + optical-flow target~\cite{bu2026laof} & 19 & 23 & 11 & 8 & 15 & 12 & 12 & 11 & 12 & 12 \\
\quad + CoMo~\cite{yang2026learning} & 16 & 14 & 18 & 8 & 14 & 16 & 14 & 12 & 14 & 14 \\
\textbf{Agent-State LAM} & \textbf{5} & \textbf{6} & \textbf{3} & \textbf{3} & \textbf{4} & \textbf{1} & \textbf{1} & \textbf{1} & \textbf{1} & \textbf{1} \\
\textit{LAOM + $32$k labels} & \textit{11} & \textit{12} & \textit{11} & \textit{8} & \textit{11} & \textit{17} & \textit{14} & \textit{12} & \textit{13} & \textit{14} \\
\textit{LAOM + $128$k labels} & \textit{9} & \textit{10} & \textit{10} & \textit{8} & \textit{9} & \textit{15} & \textit{13} & \textit{11} & \textit{13} & \textit{13} \\
\midrule
\multicolumn{11}{@{}l@{}}{LAPO, probe of its latent action ($R^2$): hop.\ \underline{79}, cheet.\ 52, walk.\ 44, hum.\ 40} \\
\bottomrule
\end{tabular}%
\hspace{6pt}%
\raisebox{-0.5\height}{\includegraphics{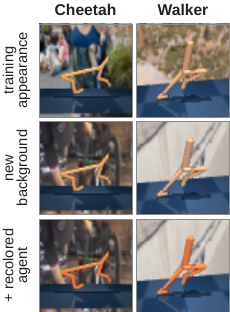}}
\end{table*}

We also check whether the baselines benefit from matching Agent-State LAM's target and loss-balancing settings: three-step prediction of unaugmented EMA targets and gradient-scale balancing of any added objective (Section~\ref{sec:exp:setup}; Section~\ref{cs:sec}).
MaskLAM has a single pixel-reconstruction loss, with no feature prediction and no added objective, so for it we instead match LAOM's residual head designs and learning-rate schedule, which the LAOM-based baselines and Agent-State LAM already share.
Across all four DCS tasks (Table~\ref{cs:tab:matched}), each baseline achieves lower action error in its original configuration, so Table~\ref{tab:dcs} reports every baseline in the configuration that is stronger \mbox{for it.}

On real robot video, we similarly examine the effects of multi-horizon prediction and learning-rate settings on fine-tuned UniVLA (Section~\ref{cs:video}).
On Bridge, using a learning rate of $10^{-4}$ for all blocks yields errors of $0.76$/$0.60$ ($z_q$/$z$), compared with $0.75$/$0.59$ in Table~\ref{tab:video}.
Using three prediction targets yields errors of $0.73$/$0.51$; neither setting closes the gap to Agent-State LAM \mbox{($0.50$/$0.33$).}

\section{Representation analysis}
\label{cs:sec:rep}
We further analyze how Agent-State LAM captures agent motion and state while reducing dependence on visual context.
We detail the controlled-intervention metrics, test robustness under natural re-renderings, analyze spatial matching in SMTD, and evaluate arm-state retrieval on \mbox{real robot video.}
\subsection{Controlled-intervention metrics}
\label{cs:interv}
Table~\ref{tab:dcs-change} of the main paper reports one measure for each of the four factors of Figure~\ref{fig:visual_context}.
Pose retrieval~\cone{} and the change under recoloring~\cfour{} are measured on re-renderings (Section~\ref{cs:natural}); the background-motion share~\ctwo{} and the change under a static swap~\cthree{} are measured on edited frames, which this subsection defines.
For these two measures, each task contributes $3{,}200$ evaluation transitions from trajectories carrying exact simulator masks.
Masks construct the edits and are not supplied to \mbox{Agent-State LAM.}

To suppress background motion, we replace the second frame's pixels outside the union of the two agent masks with the corresponding first-frame pixels.
Pixels inside that union are preserved, including any newly revealed background there.
The complementary edit suppresses agent motion by replacing the second frame's pixels inside the union with those of the first frame.
With $d_{\mathrm{bg}}$ and $d_{\mathrm{agent}}$ denoting the latent changes under freezing background and agent motion, respectively, the background's share of the motion response is \mbox{$d_{\mathrm{bg}}/(d_{\mathrm{bg}}+d_{\mathrm{agent}})$.}

For the static-swap measure, each transition is paired with a transition of another trajectory at the same time step.
Pixels that lie outside the agent masks of both transitions and change by at most $8/255$ between the two frames of each transition are replaced, in both frames, by the partner's pixels.
For each edit, the latent change is the mean squared distance between the latents of the edited and unedited transitions, normalized by the mean squared distance between two random transitions; the motion share and the static-swap change are then averaged over the \mbox{four tasks.}

Figure~\ref{fig:rho} (right) repeats the background-motion share at higher distractor speeds.
Agent-State LAM and LAOM are trained on hopper with the background video advanced by $2$, $4$, or $16$ clip frames per environment step and tested on unseen agent trajectories with unseen videos at the same speed; the $1\times$ bars evaluate the models of Table~\ref{tab:dcs}.
The faster runs train on $5{,}088$ trajectories that we collected and rendered at each speed, the $1\times$ models on the $5{,}000$ released ones, and both methods share the training data at each speed; the share is measured on \mbox{$3{,}200$ transitions per speed.}

\subsection{Natural re-rendering tests}
\label{cs:natural}
The introduction quotes three measurements of how existing latent action models depend on visual context: a linear probe recovers up to $79\%$ of the background's change pattern from LAPO's latent, replacing the background changes LAOM's features by up to $38\%$, and recoloring the agent changes MaskLAM's representation by $9\%$ on average and by up to $20\%$.
This subsection gives the test behind these numbers; Table~\ref{cs:tab:natural} holds its results for every method and the probe of LAPO, with the four numbers underlined.
No pixel is edited in this test: the simulator renders the same held-out agent trajectories, with the same camera path, at the training appearance, with another background video, and with that video and a recolored agent (examples next to Table~\ref{cs:tab:natural}), so the agent's physical state is identical in the renderings that are compared.
We use $100$ held-out trajectories on hopper and $30$ on each other task, with $32$ frames each, and recolor the agent to RGB $(0.8,0.4,0.2)$ \mbox{in $30$ trajectories per task.}

Table~\ref{cs:tab:natural} reports, in percent, how much each method's per-frame representation changes between two renderings of the same frame: the mean squared change, divided by the mean squared distance between random frames of different trajectories.
The change under a new background video compares the first two renderings, and the change under recoloring compares the last two, in which only the agent's color differs.
The representation is LAOM's encoder output for LAOM and its variants, the latent of a frame paired with itself for LAPO and MaskLAM, which have no single-frame encoder because their inverse models take a pair of frames, and the compact state for Agent-State LAM.
With another background video, LAOM's representation changes by up to $38\%$, on walker, whereas Agent-State LAM's changes by $3\%$ to $6\%$, the smallest change on every task.
Recoloring the agent changes MaskLAM's representation by $9\%$ on average and by $20\%$ on cheetah, and Agent-State LAM's by $1\%$ on every task.
LAPO's representation also changes little under recoloring, but it is largely determined by the background video: it changes by $81\%$ on average under a new background, and its pose retrieval rate is $18\%$ (Table~\ref{cs:tab:poseret}).
Agent-State LAM has the lowest mean change under both re-renderings, below LAOM trained with $128$k action labels on every task ($4\%$ and $1\%$ against $9\%$ and $13\%$ on average), together with the highest pose retrieval rate.
The task means of the recolored-agent block of Table~\ref{cs:tab:natural} are measure~4 of \mbox{Table~\ref{tab:dcs-change}.}

The first of the three measurements comes from a linear probe of LAPO's latent action.
Its target is the spatial pattern of the background change: the gray-level difference between $o_t$ and $o_{t+1}$ with the agent's pixels in either frame set to zero, averaged over an $8\times8$ grid of cells.
A ridge regression from the latent to these $64$ targets is evaluated with five folds by trajectory on the frame pairs rendered at the training appearance ($3{,}200$ on hopper and $960$ on each other task), with its penalty chosen on held-out training trajectories and $R^2$ pooled over the standardized targets.
It recovers $79\%$ / $52\%$ / $44\%$ / $40\%$ of the pattern's variance on hopper / cheetah / walker / humanoid, the last \mbox{row of Table~\ref{cs:tab:natural}.}

\begin{table}[t]
\centering
\setlength{\tabcolsep}{5pt}
\caption{\textbf{Pose retrieval across re-renderings} (mean over the four DCS tasks). R@1 (\%) $\uparrow$, measure~1 of Table~\ref{tab:dcs-change}: share of frames whose nearest frame, in a re-rendering with a new background video, shows one of the $1\%$ closest poses. In the second column, the re-rendering also changes the agent's limb thickness. Bold: best; italic: \mbox{label-supervised references.}}
\label{cs:tab:poseret}
\vspace{3pt}
\begin{tabular*}{\columnwidth}{@{\extracolsep{\fill}}l cc@{}}
\toprule
 & new & + new \\
 & background & thickness \\
\midrule
LAPO~\cite{schmidt2024learning} & 18 & 16 \\
MaskLAM~\cite{fechner2026segment} & 49 & 47 \\
LAOM~\cite{nikulin2025latent} & 40 & 38 \\
\quad + Seq-$\Delta$REPA~\cite{jiang2026olaf} & 37 & 34 \\
\quad + optical-flow target~\cite{bu2026laof} & 46 & 43 \\
\quad + CoMo~\cite{yang2026learning} & 44 & 41 \\
\textbf{Agent-State LAM} & \textbf{63} & \textbf{58} \\
\textit{LAOM + $32$k labels} & \textit{55} & \textit{51} \\
\textit{LAOM + $128$k labels} & \textit{56} & \textit{52} \\
\bottomrule
\end{tabular*}
\end{table}

Table~\ref{tab:dcs-change} also measures pose retrieval across a new background video, and Figure~\ref{fig:dcs-change} shows examples in which the agent's appearance changes as well; Table~\ref{cs:tab:poseret} gives the retrieval rate under a new background and when the limb thickness also changes.
For each task, we take $32$ frames from each of $30$ held-out trajectories and retrieve, for each frame, the nearest representation by cosine similarity among the frames of the other trajectories, re-rendered with a new background video.
In the second case, the re-rendering also scales the limb radii by $0.6$ or $1.5$; each case is averaged over two re-renderings.
A retrieval is correct when the returned frame's pose, the positions of the agent's body parts in the simulator, is among the $1\%$ closest to the query's pose (chance $1\%$).
Agent-State LAM retrieves a correct pose for $63\%$ of the frames under a new background and $58\%$ when the limb thickness also changes, the highest rate in both cases and above \mbox{LAOM trained with action labels.}

\subsection{Spatial matching in SMTD}
\label{cs:smtd}
Section~\ref{sec:method:smtd} states that matching across locations roughly halves the measured change in distractor patches and leaves it below the agent's change in nine of ten trajectories.
This subsection gives the measurement behind \mbox{that statement.}

SMTD compares each state with frozen patch features of nearby frames, so we measure how much the agent and the distractor change between two frames as these features see them.
We use the frozen DINOv2 ViT-L/14 patch tokens of $64$ DCS hopper episodes, with the simulator's agent masks for this measurement only: a patch counts as an agent patch if more than half of its pixels belong to the agent, and as a distractor patch if fewer than $5\%$ do.
The change of a patch is one minus the cosine similarity between its token and a token of the later frame: the token at the same location, or, mirroring the spatially flexible matching of Eq.~\eqref{eq:method:score}, the best-matching token anywhere in that frame.
Matching changes the distractor more than the agent.
At gaps of one to eight steps, it lowers the mean change of distractor patches by $43$--$50\%$ (from $0.248$ to $0.141$ at one step) and that of agent patches by $12$--$32\%$ (from $0.204$ to $0.180$), because background content that shifts or pans is found again elsewhere in the later frame, whereas a change of pose alters the agent's patches themselves.

\begin{table}[!ht]
\centering
\setlength{\tabcolsep}{6pt}
\caption{\textbf{Spatial matching on DCS hopper.} Number of episodes, out of $64$, in which the agent's patches change more than the distractor's between two frames ($\uparrow$), for frames $1$ to $8$ steps apart. The change of a patch is one minus the cosine similarity of frozen DINOv2 features, either with the patch at the same location in the later frame or with its best match anywhere in that frame, mirroring the spatially flexible matching of Eq.~\eqref{eq:method:score}. With matching, the agent changes more than the distractor \mbox{in nine of ten episodes.}}
\label{cs:tab:smtd-match}
\vspace{3pt}
\begin{tabular}{@{}lccccc@{}}
\toprule
gap in steps & 1 & 2 & 3 & 4 & 8 \\
\midrule
same location & 12 & 13 & 13 & 15 & 24 \\
best match anywhere (SMTD) & \textbf{59} & \textbf{58} & \textbf{59} & \textbf{58} & \textbf{59} \\
\bottomrule
\end{tabular}
\end{table}

The ordering of the two changes therefore flips (Table~\ref{cs:tab:smtd-match}).
At the same location, the agent's change exceeds the distractor's in $12$ to $24$ of the $64$ episodes; after matching, it does so in $58$ or $59$, \mbox{nine of ten.}

\begin{table}[!ht]
\centering
\setlength{\tabcolsep}{6pt}
\caption{\textbf{Where the trained state matches on DCS hopper.} On $2{,}048$ held-out frames: share of frames in which the patch that best matches the model's state touches the agent, and share of the patch weights of the smooth maximum in Eq.~\eqref{eq:method:score} that lies on agent patches (\%, $\uparrow$). The agent touches $7\%$ of the patches of a frame. The first two rows are \mbox{SMTD variants of Table~\ref{tab:ablation}.}}
\label{cs:tab:smtd-hit}
\vspace{3pt}
\begin{tabular}{@{}lcc@{}}
\toprule
 & best match & weight \\
 & on agent $\uparrow$ & on agent $\uparrow$ \\
\midrule
positives only & 1 & 2 \\
frame-mean scoring & 19 & 13 \\
\textbf{Agent-State LAM} & \textbf{93} & \textbf{89} \\
\bottomrule
\end{tabular}
\end{table}
\begin{table*}[t]
\centering
\setlength{\tabcolsep}{5pt}
\caption{\textbf{Arm-state retrieval on real robot video.} R@1 (\%) $\uparrow$: share of $8{,}000$ held-out frames whose nearest frame from another episode, by cosine similarity of each model's per-frame vector (Section~\ref{cs:rvret}), shows the same recorded arm state: the end effector among the $1\%$ closest positions and the gripper in the same open or closed state (Jaco and Taco record no binary open/closed gripper state, so position alone is used). The first row scores the frozen DINOv2-B patch tokens that every model receives, averaged over the frame. Figure~\ref{cs:fig:rv-retrieval} shows one query per dataset. Bold: best.}
\label{cs:tab:rv-retrieval}
\vspace{3pt}
\begin{tabular}{@{}l cccccc@{}}
\toprule
 & Bridge & RT-1 & UR5 & IRASim & Jaco & Taco \\
\midrule
\textit{DINOv2-B~\cite{oquab2023dinov2} features, frozen} & \textit{12.3} & \textit{9.0} & \textit{54.0} & \textit{16.3} & \textit{28.9} & \textit{43.3} \\
UniVLA~\cite{bu2025univla}, released & 17.2 & 16.3 & 39.9 & 19.8 & 34.3 & 35.2 \\
UniVLA, fine-tuned & 16.3 & 17.2 & 43.2 & 18.8 & 38.1 & 36.0 \\
\quad + Seq-$\Delta$REPA~\cite{jiang2026olaf} & 21.3 & 20.5 & 54.9 & 24.2 & 49.5 & 44.3 \\
\quad + optical-flow target~\cite{bu2026laof} & 18.3 & 22.7 & 48.0 & 21.2 & 42.2 & 40.5 \\
\quad + CoMo~\cite{yang2026learning} & 16.1 & 18.1 & 42.7 & 19.4 & 40.8 & 35.5 \\
\quad + MaskLAM~\cite{fechner2026segment} & 16.8 & 18.1 & 42.9 & 20.3 & 43.6 & 36.1 \\
\quad \textbf{+ Agent-State LAM} & \textbf{22.7} & \textbf{35.0} & \textbf{56.3} & \textbf{33.8} & \textbf{50.2} & \textbf{50.6} \\
\qquad without SMTD & 19.9 & 26.1 & 48.5 & 21.4 & 28.4 & 46.7 \\
\bottomrule
\end{tabular}
\end{table*}

\begin{figure*}[t]
\centering
\includegraphics[width=0.95\textwidth]{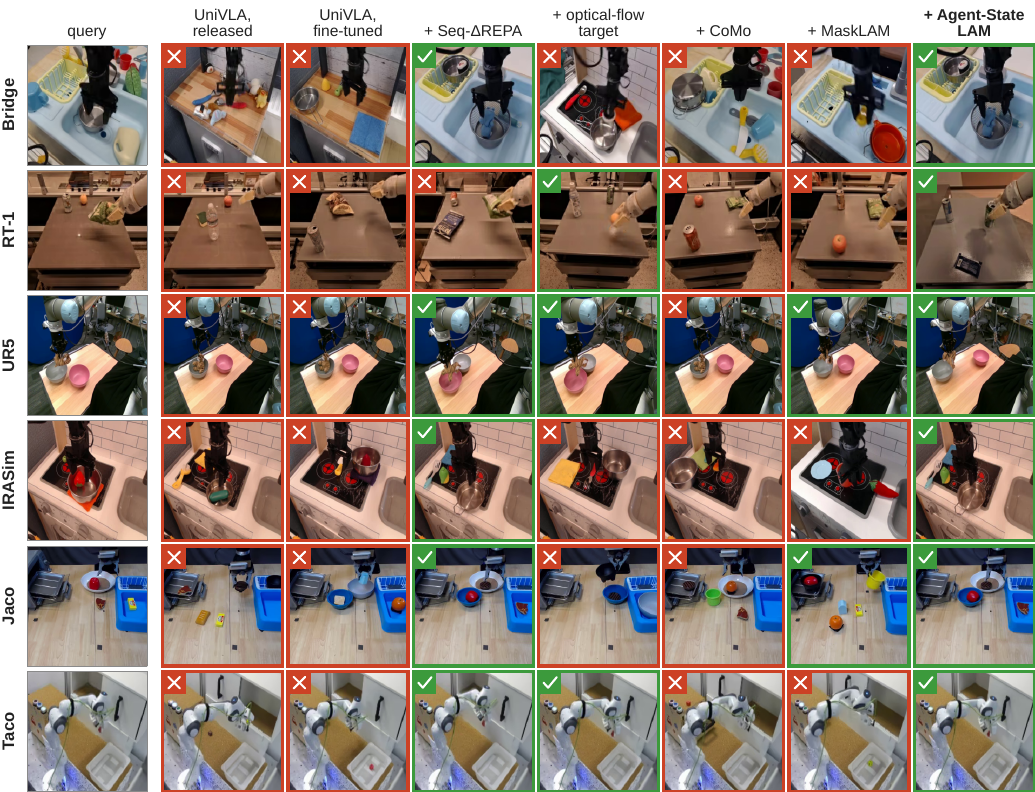}
\caption{\textbf{Arm-state retrieval on real robot video: one query per dataset.} Each row shows a query frame (left) and the nearest-neighbor frame from another episode returned by each model under the protocol of Table~\ref{cs:tab:rv-retrieval}. A green frame with a check marks a return that shows the same recorded arm state as the query; a red frame with a cross marks a return that \mbox{does not.}}
\label{cs:fig:rv-retrieval}
\end{figure*}

The trained model uses this signal (Table~\ref{cs:tab:smtd-hit}).
The patch that best matches Agent-State LAM's state touches the agent in $93\%$ of the frames, and $89\%$ of the smooth maximum's patch weight lies on agent patches, although the agent touches only $7\%$ of the patches of a frame; with positives only, these shares are $1\%$ and $2\%$, and with frame-mean scoring, \mbox{$19\%$ and $13\%$.}

\subsection{Arm-state retrieval on real robot video}
\label{cs:rvret}
Pose retrieval on DCS (Table~\ref{tab:dcs-change}) asks whether frames with similar agent poses lie close together in a model's representation.
The robot datasets record the state of the arm, so we run the corresponding test on real robot video (Table~\ref{cs:tab:rv-retrieval}).

From the held-out episodes of each dataset, we draw $8{,}000$ frames in which the arm is visible.
Each frame is a query, and its nearest frame from another episode is retrieved by cosine similarity.
A retrieval is correct if the returned frame shows the same recorded arm state: its end effector is among the $1\%$ closest positions and, on the four datasets that record a binary open/closed gripper state, the gripper is in the same state.
Agent-State LAM is represented by its compact state.
The UniVLA-based models have no per-frame state; to evaluate them favorably, each is represented by the better of two vectors that its encoder computes from the frame alone: its latent-action tokens and the mean of its output patch tokens.
As a reference, the first row of the table scores the mean of the frozen DINOv2-B patch tokens that \mbox{every model receives.}

Agent-State LAM has the highest retrieval rate on all six datasets, and Figure~\ref{cs:fig:rv-retrieval} shows one query per dataset with the frame that every model returns for it.
On RT-1, its state returns the same arm state for $35.0\%$ of the queries, against $22.7\%$ for the optical-flow-target variant; on IRASim and Taco, it does so for $33.8\%$ and $50.6\%$, against $24.2\%$ and $44.3\%$ for Seq-$\Delta$REPA.
Relative to the frozen DINOv2-B input features, Agent-State LAM improves retrieval R@1 on all six datasets, by up to $+26.0$ points on RT-1 ($35.0\%$ vs.\ $9.0\%$) and $+21.3$ on Jaco ($50.2\%$ vs.\ $28.9\%$).
SMTD contributes to this organization of the state: without it, the retrieval rate is lower on every dataset, by $3$ to $22$ points, with the largest drops on Jaco ($50.2\%$ to $28.4\%$), IRASim ($33.8\%$ to $21.4\%$), and RT-1 ($35.0\%$ to $26.1\%$), so the compact state alone does not \mbox{account for it.}

\makeatletter
\newcommand{\cswmtables}{%
  \begin{table*}[t]
\centering
\setlength{\tabcolsep}{2.3pt}
\caption{\textbf{World-model prediction from decoded latent actions.} A linear decoder fitted on training data maps each model's quantized latents to robot commands, which drive a frozen generator from the real first frame: Cosmos-Predict2 on $200$ IRASim clips and Cosmos~3 on $400$ Bridge and $200$ RT-1 clips. PSNR: generated versus real continuation. $R^2$, $R^2_{\mathrm{rot}}$: translation and rotation recovered from the real test videos, pooled over clips and steps and averaged over components. $R^2_{\mathrm{read}}$: translation recovered from videos generated with the true commands. Traj.: WorldArena trajectory accuracy~\cite{shang2026worldarena} of the tracked arm (Section~\ref{cs:wm-eval}). The three $R^2$ columns apply only to latent action models. Bold: best latent action model.}
\label{cs:tab:wm-full}
\vspace{3pt}
\begin{tabular}{@{}l ccccc ccccc ccccc@{}}
\toprule
 & \multicolumn{5}{c}{IRASim, Cosmos-Predict2} & \multicolumn{5}{c}{Bridge, Cosmos 3} & \multicolumn{5}{c}{RT-1, Cosmos 3} \\
\cmidrule(lr){2-6} \cmidrule(lr){7-11} \cmidrule(lr){12-16}
conditioning & PSNR & $R^2$ & $R^2_{\mathrm{rot}}$ & $R^2_{\mathrm{read}}$ & Traj.\ & PSNR & $R^2$ & $R^2_{\mathrm{rot}}$ & $R^2_{\mathrm{read}}$ & Traj.\ & PSNR & $R^2$ & $R^2_{\mathrm{rot}}$ & $R^2_{\mathrm{read}}$ & Traj.\ \\
 & $\uparrow$ & $\uparrow$ & $\uparrow$ & $\uparrow$ & $\uparrow$ & $\uparrow$ & $\uparrow$ & $\uparrow$ & $\uparrow$ & $\uparrow$ & $\uparrow$ & $\uparrow$ & $\uparrow$ & $\uparrow$ & $\uparrow$ \\[-2pt]
\midrule
\textit{true actions (upper)} & \textit{26.05} &  &   &  & \textit{97.1} & \textit{19.73} &  &   &  & \textit{65.6} & \textit{23.73} &  &   &  & \textit{86.9} \\
\textit{other clip's actions (lower)} & \textit{21.46} &  &   &  & \textit{31.0} & \textit{17.05} &  &   &  & \textit{22.0} & \textit{19.51} &  &   &  & \textit{21.6} \\
\midrule
UniVLA, released & 21.96 & 0.05 & $-0.01$ & 0.00 & 41.9 & 18.15 & 0.13 & 0.01 & 0.09 & 37.8 & 21.10 & 0.16 & 0.01 & 0.10 & 47.9 \\
UniVLA, fine-tuned & 22.38 & 0.14 & 0.00 & 0.09 & 51.1 & 18.24 & 0.15 & 0.02 & 0.10 & 41.6 & 21.18 & 0.23 & 0.02 & 0.20 & 48.2 \\
\quad + Seq-$\Delta$REPA~\cite{jiang2026olaf} & 22.43 & 0.15 & 0.00 & 0.13 & 51.2 & 18.44 & 0.21 & \textbf{0.03} & 0.19 & 44.9 & 20.99 & 0.18 & 0.02 & 0.15 & 47.1 \\
\quad + optical-flow target~\cite{bu2026laof} & 21.89 & 0.06 & \textbf{0.01} & 0.03 & 39.5 & 18.50 & 0.18 & \textbf{0.03} & 0.15 & 42.6 & 21.09 & 0.37 & 0.07 & 0.33 & \textbf{56.4} \\
\quad + CoMo~\cite{yang2026learning} & 22.63 & 0.20 & \textbf{0.01} & 0.17 & 56.8 & 18.49 & 0.16 & 0.02 & 0.13 & 43.0 & 21.20 & 0.22 & 0.00 & 0.18 & 51.2 \\
\quad + MaskLAM~\cite{fechner2026segment} & 22.45 & 0.14 & 0.00 & 0.08 & 53.2 & 18.17 & 0.13 & 0.02 & 0.00 & 39.7 & 21.12 & 0.25 & 0.02 & 0.20 & 48.6 \\
\quad \textbf{+ Agent-State LAM} & \textbf{23.35} & \textbf{0.41} & \textbf{0.01} & \textbf{0.33} & \textbf{67.5} & \textbf{18.63} & \textbf{0.34} & \textbf{0.03} & \textbf{0.25} & \textbf{45.8} & \textbf{21.37} & \textbf{0.48} & \textbf{0.14} & \textbf{0.46} & 53.3 \\
\bottomrule
\end{tabular}
\end{table*}
  \begin{table*}[t]
\centering
\setlength{\tabcolsep}{7pt}
\caption{\textbf{Motion transfer across visual contexts.} For randomly paired test clips (Section~\ref{cs:wm-pred}), commands decoded from video A drive the frozen generator from the first frame of video B. Both measures compare the generated video with the reference generated from the same first frame and random seed using the true actions of A: WorldArena trajectory accuracy of the tracked arm (Traj.) and PSNR. Bold: best per column.}
\label{cs:tab:wm-transfer}
\vspace{3pt}
\begin{tabular}{@{}l cc cc cc@{}}
\toprule
 & \multicolumn{2}{c}{IRASim, Cosmos-Predict2} & \multicolumn{2}{c}{Bridge, Cosmos~3} & \multicolumn{2}{c}{RT-1, Cosmos~3} \\
\cmidrule(lr){2-3} \cmidrule(lr){4-5} \cmidrule(l){6-7}
conditioning & Traj.\ $\uparrow$ & PSNR $\uparrow$ & Traj.\ $\uparrow$ & PSNR $\uparrow$ & Traj.\ $\uparrow$ & PSNR $\uparrow$ \\
\midrule
UniVLA, released & 45.0 & 22.64 & 39.9 & 18.68 & 41.6 & 21.47 \\
UniVLA, fine-tuned & 52.8 & 23.37 & 40.8 & 18.77 & 43.6 & 21.52 \\
\quad + Seq-$\Delta$REPA & 52.7 & 23.27 & 43.2 & 19.04 & 41.8 & 21.39 \\
\quad + optical-flow target & 41.5 & 22.78 & 42.0 & 19.16 & \textbf{46.8} & 21.48 \\
\quad + CoMo & 54.3 & 23.84 & 43.6 & 19.19 & 41.9 & 21.59 \\
\quad + MaskLAM & 51.7 & 23.49 & 39.9 & 18.73 & 41.8 & 21.49 \\
\quad \textbf{+ Agent-State LAM} & \textbf{66.8} & \textbf{24.98} & \textbf{45.2} & \textbf{19.40} & 45.4 & \textbf{21.81} \\
\bottomrule
\end{tabular}
\end{table*}
}
\makeatother
\section{Further ablations and multi-seed results}
\label{cs:sec:abl}
We provide additional ablations to examine the contributions of Agent-State LAM's components and training objectives.
We also report results across multiple training seeds to assess the consistency of the observed improvements.
\subsection{Component ablation on Bridge}
\label{cs:abl-video}
Table~\ref{tab:ablation} of the main paper ablates the compact state and SMTD on DCS hopper, and Table~\ref{cs:tab:abl-bridge} adds the two components to the fine-tuned UniVLA on Bridge one at a time.
The compact state is important on real video.
UniVLA's latent action model works on $256$ DINOv2-B patch tokens per frame, which are high-dimensional and can retain much visual content that is unrelated to the action.
With the compact state in place of these tokens as the inverse model's input, the probe error falls from $0.75$ / $0.59$ ($z_q$ / $z$) to $0.63$ / $0.45$, and with the compact state as the prediction target as well it falls to $0.55$ / $0.37$.
SMTD lowers the error further, from $0.55$ to $0.50$ on the quantized latent and from $0.37$ to $0.33$ on the continuous latent; without SMTD, the arm-state retrieval rate also drops on all six real-video datasets, by $3$ to $22$ points \mbox{(Table~\ref{cs:tab:rv-retrieval}).}

\begin{table}[!ht]
\centering
\setlength{\tabcolsep}{4pt}
\caption{\textbf{Ablation on Bridge.} Normalized probe error $\downarrow$ on Bridge (Table~\ref{tab:video} protocol, step $10{,}000$). Each row adds one component to the row above it; the last two rows differ \mbox{only in SMTD.}}
\label{cs:tab:abl-bridge}
\vspace{3pt}
\begin{tabular*}{\columnwidth}{@{\extracolsep{\fill}}l ccc@{}}
\toprule
 & SMTD & $z_q$ & $z$ \\
\midrule
UniVLA, fine-tuned & $\times$ & 0.75 & 0.59 \\
\quad + compact state as input & $\times$ & 0.63 & 0.45 \\
\quad + compact state as target & $\times$ & 0.55 & 0.37 \\
\quad \textbf{+ SMTD (Agent-State LAM)} & \checkmark & \textbf{0.50} & \textbf{0.33} \\
\bottomrule
\end{tabular*}
\end{table}

\subsection{Ablation variants on DCS}
\label{cs:abl}
For completeness, we summarize the implementation of the Table~\ref{tab:ablation} ablation variants on DCS hopper, each changing one component of Agent-State LAM under the same training budget.
In \emph{without compact state}, the inverse model, forward model, and prediction targets operate on dense $8\times8\times192$ backbone features, while SMTD is applied to the backbone through an auxiliary attention-pooling readout head that feeds only $\mathcal{L}_{\mathrm{SMTD}}$; pose R@1 is evaluated on the dense encoder features read by the inverse model (as for LAOM and \emph{dense IDM input}, where only the inverse model reads dense features while forward prediction and SMTD retain the compact state).
In \emph{agent mask input}, SMTD is turned off and, at both training and test time, every frame keeps only the pixels inside its SAM~2.1 agent mask (Section~\ref{cs:base}), with all other pixels set \mbox{to gray.}

Among the SMTD variants, \emph{positives only} omits the negative contrasts in Eq.~\eqref{eq:method:smtd} and directly minimizes $-\mathbb{E}_{\mathcal{V},t}[r_{t,t}/\tau]$; \emph{frame-mean scoring} replaces the smooth maximum over patches in Eq.~\eqref{eq:method:score} with the mean of the patch similarities, $r_{t,u}=\frac{1}{N_p}\sum_{p}\cos(q_t,f_{u,p})$, which scores the state against the mean frozen DINOv2-L patch feature of frame $u$; \emph{state-to-state contrast} replaces the frozen patch features of each candidate frame with the model's own online compact state of that frame, scored by cosine similarity in Eq.~\eqref{eq:method:smtd} with the same temperature $\tau$; and \emph{DINOv2-S teacher} takes the frozen patch features from DINOv2-S instead of \mbox{DINOv2-L.}

While Table~\ref{tab:ablation} compares SMTD against its own architectural and contrastive variants on DCS hopper, Table~\ref{cs:tab:abl-dcs} tests replacing SMTD on the compact state with two alternative auxiliary objectives.
Neither substitutes for SMTD.
Replacing SMTD with the masked optical-flow target lowers action error only modestly relative to omitting SMTD ($2.68$ vs.\ $3.32$) and leaves pose retrieval low ($14\%$ vs.\ $10\%$, compared with $1.14$ and $67\%$ with SMTD), suggesting that flow targets supervise motion without organizing a context-invariant per-frame state.
Directly regressing the compact state onto the mean frozen DINOv2-L features of the frame degrades performance sharply to $9.83$ action error and $4\%$ pose R@1---indicating that access to frozen DINOv2-L features alone does not explain Agent-State LAM's gains, as global frame features remain dominated by \mbox{the background video.}

\begin{table}[!ht]
\centering
\setlength{\tabcolsep}{5pt}
\caption{\textbf{Replacing SMTD with alternative objectives} (DCS hopper; metrics as in Table~\ref{tab:ablation}, from which the first two rows are taken). Optical-flow target: the masked optical-flow target of the optical-flow baseline (using the SAM~2.1 agent masks of MaskLAM); feature regression: cosine regression of the compact state onto the mean frozen DINOv2-L patch feature \mbox{of its frame.}}
\label{cs:tab:abl-dcs}
\vspace{3pt}
\begin{tabular*}{\columnwidth}{@{\extracolsep{\fill}}l cc@{}}
\toprule
 & action & pose \\
 & error $\downarrow$ & R@1 $\uparrow$ \\
\midrule
\textbf{Agent-State LAM} & \textbf{1.14} & \textbf{67} \\
\hspace{0.6em}without SMTD & 3.32 & 10 \\
\hspace{0.6em}optical-flow target in place of SMTD & 2.68 & 14 \\
\hspace{0.6em}feature regression in place of SMTD & 9.83 & 4 \\
\bottomrule
\end{tabular*}
\end{table}

\subsection{Results across training seeds}
\label{cs:seeds}
Table~\ref{tab:dcs} reports a single training run per cell (seed~$0$).
Following Section~\ref{sec:exp:dcs}, Table~\ref{cs:tab:dcs-seeds} adds two independent training seeds (seeds~$1$ and~$2$) for both Agent-State LAM and LAOM across all four DCS tasks and all three stages.
Each Stage-1 checkpoint receives its own cloned policy and decoders while holding the policy-stage seed and evaluation episodes fixed, isolating variability from Stage-1 LAM training.
Across every task, Agent-State LAM's highest Stage-1 action error remains below LAOM's lowest seed, and in every task and control stage its lowest normalized return exceeds LAOM's highest seed---with zero overlap across all twelve task--stage cells, indicating that the improvements over LAOM are consistent across \mbox{training runs.}

\begin{table}[!ht]
\centering
\setlength{\tabcolsep}{0pt}%
\caption{\textbf{Multi-seed results on DCS.} Action error $\downarrow$ (Stage~1) and normalized return $\uparrow$ (Stages~2--3) under the Table~\ref{tab:dcs} protocol: mean $\pm$ sample standard deviation over three independent training seeds (seed~$0$ is the run reported in Table~\ref{tab:dcs}). Bold: better mean.}
\label{cs:tab:dcs-seeds}
\vspace{3pt}
\begin{tabular*}{\columnwidth}{@{\extracolsep{\fill}}lcccc@{}}
\toprule
 & hopper & cheetah & walker & humanoid \\
\midrule
\multicolumn{5}{@{}l}{\textit{LAOM}} \\
Stage 1 & $2.60{\pm}0.13$ & $2.67{\pm}0.17$ & $6.04{\pm}1.03$ & $0.147{\pm}0.001$ \\
Stage 2 & $0.22{\pm}0.06$ & $0.43{\pm}0.14$ & $0.40{\pm}0.08$ & $0.10{\pm}0.08$ \\
Stage 3 & $0.11{\pm}0.02$ & $0.43{\pm}0.02$ & $0.08{\pm}0.03$ & $0.028{\pm}0.005$ \\
\midrule
\multicolumn{5}{@{}l}{\textit{Agent-State LAM}} \\
Stage 1 & $\mathbf{1.12}{\pm}0.02$ & $\mathbf{1.06}{\pm}0.12$ & $\mathbf{2.37}{\pm}0.10$ & $\mathbf{0.112}{\pm}0.001$ \\
Stage 2 & $\mathbf{0.65}{\pm}0.15$ & $\mathbf{0.93}{\pm}0.06$ & $\mathbf{0.81}{\pm}0.05$ & $\mathbf{0.27}{\pm}0.09$ \\
Stage 3 & $\mathbf{0.36}{\pm}0.01$ & $\mathbf{0.66}{\pm}0.01$ & $\mathbf{0.28}{\pm}0.07$ & $\mathbf{0.056}{\pm}0.001$ \\
\bottomrule
\end{tabular*}
\end{table}
  \begin{table*}[t]
\centering
\setlength{\tabcolsep}{2.3pt}
\caption{\textbf{World-model prediction from decoded latent actions.} A linear decoder fitted on training data maps each model's quantized latents to robot commands, which drive a frozen generator from the real first frame: Cosmos-Predict2 on $200$ IRASim clips and Cosmos~3 on $400$ Bridge and $200$ RT-1 clips. PSNR: generated versus real continuation. $R^2$, $R^2_{\mathrm{rot}}$: translation and rotation recovered from the real test videos, pooled over clips and steps and averaged over components. $R^2_{\mathrm{read}}$: translation recovered from videos generated with the true commands. Traj.: WorldArena trajectory accuracy~\cite{shang2026worldarena} of the tracked arm (Section~\ref{cs:wm-eval}). The three $R^2$ columns apply only to latent action models. Bold: best latent action model.}
\label{cs:tab:wm-full}
\vspace{3pt}
\begin{tabular}{@{}l ccccc ccccc ccccc@{}}
\toprule
 & \multicolumn{5}{c}{IRASim, Cosmos-Predict2} & \multicolumn{5}{c}{Bridge, Cosmos 3} & \multicolumn{5}{c}{RT-1, Cosmos 3} \\
\cmidrule(lr){2-6} \cmidrule(lr){7-11} \cmidrule(lr){12-16}
conditioning & PSNR & $R^2$ & $R^2_{\mathrm{rot}}$ & $R^2_{\mathrm{read}}$ & Traj.\ & PSNR & $R^2$ & $R^2_{\mathrm{rot}}$ & $R^2_{\mathrm{read}}$ & Traj.\ & PSNR & $R^2$ & $R^2_{\mathrm{rot}}$ & $R^2_{\mathrm{read}}$ & Traj.\ \\
 & $\uparrow$ & $\uparrow$ & $\uparrow$ & $\uparrow$ & $\uparrow$ & $\uparrow$ & $\uparrow$ & $\uparrow$ & $\uparrow$ & $\uparrow$ & $\uparrow$ & $\uparrow$ & $\uparrow$ & $\uparrow$ & $\uparrow$ \\[-2pt]
\midrule
\textit{true actions (upper)} & \textit{26.05} &  &   &  & \textit{97.1} & \textit{19.73} &  &   &  & \textit{65.6} & \textit{23.73} &  &   &  & \textit{86.9} \\
\textit{other clip's actions (lower)} & \textit{21.46} &  &   &  & \textit{31.0} & \textit{17.05} &  &   &  & \textit{22.0} & \textit{19.51} &  &   &  & \textit{21.6} \\
\midrule
UniVLA, released & 21.96 & 0.05 & $-0.01$ & 0.00 & 41.9 & 18.15 & 0.13 & 0.01 & 0.09 & 37.8 & 21.10 & 0.16 & 0.01 & 0.10 & 47.9 \\
UniVLA, fine-tuned & 22.38 & 0.14 & 0.00 & 0.09 & 51.1 & 18.24 & 0.15 & 0.02 & 0.10 & 41.6 & 21.18 & 0.23 & 0.02 & 0.20 & 48.2 \\
\quad + Seq-$\Delta$REPA~\cite{jiang2026olaf} & 22.43 & 0.15 & 0.00 & 0.13 & 51.2 & 18.44 & 0.21 & \textbf{0.03} & 0.19 & 44.9 & 20.99 & 0.18 & 0.02 & 0.15 & 47.1 \\
\quad + optical-flow target~\cite{bu2026laof} & 21.89 & 0.06 & \textbf{0.01} & 0.03 & 39.5 & 18.50 & 0.18 & \textbf{0.03} & 0.15 & 42.6 & 21.09 & 0.37 & 0.07 & 0.33 & \textbf{56.4} \\
\quad + CoMo~\cite{yang2026learning} & 22.63 & 0.20 & \textbf{0.01} & 0.17 & 56.8 & 18.49 & 0.16 & 0.02 & 0.13 & 43.0 & 21.20 & 0.22 & 0.00 & 0.18 & 51.2 \\
\quad + MaskLAM~\cite{fechner2026segment} & 22.45 & 0.14 & 0.00 & 0.08 & 53.2 & 18.17 & 0.13 & 0.02 & 0.00 & 39.7 & 21.12 & 0.25 & 0.02 & 0.20 & 48.6 \\
\quad \textbf{+ Agent-State LAM} & \textbf{23.35} & \textbf{0.41} & \textbf{0.01} & \textbf{0.33} & \textbf{67.5} & \textbf{18.63} & \textbf{0.34} & \textbf{0.03} & \textbf{0.25} & \textbf{45.8} & \textbf{21.37} & \textbf{0.48} & \textbf{0.14} & \textbf{0.46} & 53.3 \\
\bottomrule
\end{tabular}
\end{table*}
  \begin{table*}[t]
\centering
\setlength{\tabcolsep}{7pt}
\caption{\textbf{Motion transfer across visual contexts.} For randomly paired test clips (Section~\ref{cs:wm-pred}), commands decoded from video A drive the frozen generator from the first frame of video B. Both measures compare the generated video with the reference generated from the same first frame and random seed using the true actions of A: WorldArena trajectory accuracy of the tracked arm (Traj.) and PSNR. Bold: best per column.}
\label{cs:tab:wm-transfer}
\vspace{3pt}
\begin{tabular}{@{}l cc cc cc@{}}
\toprule
 & \multicolumn{2}{c}{IRASim, Cosmos-Predict2} & \multicolumn{2}{c}{Bridge, Cosmos~3} & \multicolumn{2}{c}{RT-1, Cosmos~3} \\
\cmidrule(lr){2-3} \cmidrule(lr){4-5} \cmidrule(l){6-7}
conditioning & Traj.\ $\uparrow$ & PSNR $\uparrow$ & Traj.\ $\uparrow$ & PSNR $\uparrow$ & Traj.\ $\uparrow$ & PSNR $\uparrow$ \\
\midrule
UniVLA, released & 45.0 & 22.64 & 39.9 & 18.68 & 41.6 & 21.47 \\
UniVLA, fine-tuned & 52.8 & 23.37 & 40.8 & 18.77 & 43.6 & 21.52 \\
\quad + Seq-$\Delta$REPA & 52.7 & 23.27 & 43.2 & 19.04 & 41.8 & 21.39 \\
\quad + optical-flow target & 41.5 & 22.78 & 42.0 & 19.16 & \textbf{46.8} & 21.48 \\
\quad + CoMo & 54.3 & 23.84 & 43.6 & 19.19 & 41.9 & 21.59 \\
\quad + MaskLAM & 51.7 & 23.49 & 39.9 & 18.73 & 41.8 & 21.49 \\
\quad \textbf{+ Agent-State LAM} & \textbf{66.8} & \textbf{24.98} & \textbf{45.2} & \textbf{19.40} & 45.4 & \textbf{21.81} \\
\bottomrule
\end{tabular}
\end{table*}

\section{World-model evaluation}
\label{cs:sec:wm}
Section~\ref{sec:exp:wm} tests whether commands decoded from latent actions can drive frozen video generators.
A command recovered from a latent action should move the generated robot as the true action does, both in the clip from which the latent was inferred and when the motion is transferred to another visual context.
This section details the evaluation setup, presents the full prediction and command-recovery results behind Table~\ref{tab:wm} (Table~\ref{cs:tab:wm-full}), quantitatively evaluates cross-context motion transfer across all three datasets (Table~\ref{cs:tab:wm-transfer}), and shows qualitative comparisons \mbox{across all fine-tuned models.}

\subsection{Evaluation setup}
\label{cs:wm-eval}
We use two released action-conditioned video generators without fine-tuning.
On IRASim, the $2$B-parameter Cosmos-Predict2 model~\cite{ali2025world} generates $12$ frames using $35$ sampling steps and guidance $7$; on Bridge and RT-1, the $4$B-parameter Cosmos~3 Edge model~\cite{agarwal2026cosmos} generates $16$ frames using $30$ UniPC steps and guidance $1$, without a text prompt so that generated motion is driven solely by the decoded commands.
A linear decoder, a ridge regression fitted on training data, maps each model's quantized latents to the generator's native command format: seven dimensions for Cosmos-Predict2 (translation, rotation, and gripper) and ten for Cosmos~3 (translation, a six-dimensional rotation, and gripper).
Matching each dataset's frame interval (Table~\ref{cs:tab:datasets}), on IRASim ($k=1$) the latent of each consecutive frame pair predicts that step's command, whereas on Bridge and RT-1 ($k=3$) the latent of frames $t$ and $t+3$ predicts the mean command over the three intervening steps.
Within each comparison, all rows share the generator, the first frame, and the random seed, so they differ only in the commands decoded from their latents.

We evaluate on $200$ IRASim, $400$ Bridge, and $200$ RT-1 test clips using two measures.
PSNR compares the generated frames with those of a reference video.
Because whole-frame similarity can obscure motion errors, we also report WorldArena's trajectory accuracy~\cite{shang2026worldarena}, which tracks the arm from the first frame and compares its image path with that in the reference video via dynamic time warping.
Because this measure requires a trackable arm in the initial frame, it is computed for all models on the same common subsets of clips with a detected arm in the real first frame: $143$ IRASim, $312$ Bridge, and $160$ RT-1 clips; PSNR is averaged \mbox{over all clips.}

\subsection{Prediction and motion transfer}
\label{cs:wm-pred}
We first measure same-clip prediction, the setting of Table~\ref{tab:wm}, in which a real continuation exists: the commands decoded from a clip drive the generator from that clip's own first frame, and the generated frames are compared with the real ones.
Table~\ref{cs:tab:wm-full} gives every measure behind Table~\ref{tab:wm}, with true commands and another clip's commands as the upper and lower \mbox{reference rows.}
Agent-State LAM has the highest PSNR on all three datasets ($23.35$, $18.63$, and $21.37$~dB, versus $22.63$, $18.50$, and $21.20$~dB for the next-best model).
Relative to the fine-tuned UniVLA, its PSNR gain, computed before rounding, is $0.97$~dB on IRASim, $0.39$~dB on Bridge (paired $95\%$ bootstrap interval $[0.26,0.52]$), and $0.20$~dB on RT-1 ($[0.05,0.34]$); for scale, the true commands gain $4.59$, $2.68$, and $4.22$~dB over another clip's commands.
On trajectory accuracy, Agent-State LAM likewise has the highest mean on IRASim ($67.5$ versus $56.8$ for the next-best model and $51.1$ for fine-tuned UniVLA) and Bridge ($45.8$ versus $44.9$ and $41.6$), and is above fine-tuned UniVLA on RT-1 ($53.3$ versus $48.2$), where the optical-flow-target \mbox{variant scores higher ($56.4$).}

We then consider motion transfer across visual contexts, the setting of Figure~\ref{fig:wm}a: the commands decoded from a video A drive the generator from the first frame of another video B, so the motion of A has to be reproduced in a scene from which the latent actions were not inferred.
Table~\ref{cs:tab:wm-transfer} quantifies this transfer on the three datasets.
On each dataset, we pair at random the clips with a detected arm, which gives $143$, $312$, and $160$ pairs on IRASim, Bridge, and RT-1, so that each clip provides the first frame of one pair and the motion of another.
No real video shows the motion of A in the scene of B, so the reference for each pair is the video that the generator produces from the same first frame with the true actions of A, and both measures compare each model's video with this reference.
Because the reference is itself generated, with the same generator and random seed, PSNR is higher here than against real video in Table~\ref{cs:tab:wm-full}.
Agent-State LAM achieves the highest PSNR on all three datasets ($24.98$, $19.40$, and $21.81$~dB, versus $23.84$, $19.19$, and $21.59$~dB for the next-best model) and the highest trajectory accuracy on IRASim ($66.8$ against $54.3$) and Bridge ($45.2$ against $43.6$); on RT-1, it is above fine-tuned UniVLA ($45.4$ against $43.6$), while the optical-flow-target variant scores higher ($46.8$), as it does in Table~\ref{tab:wm}.

\subsection{Command recovery analysis}
\label{cs:wm}
Table~\ref{cs:tab:wm-full} also reports how accurately the decoded commands recover the true robot actions.
Agent-State LAM achieves the highest end-effector translation recovery across all three datasets, both on real test videos ($R^2$ of $0.41$, $0.34$, and $0.48$ on IRASim, Bridge, and RT-1, versus $0.14$, $0.15$, and $0.23$ for fine-tuned UniVLA and at most $0.20$, $0.21$, and $0.37$ for any baseline) and when read out from generated true-action rollouts ($R^2_{\mathrm{read}}$ of $0.33$, $0.25$, and $0.46$, versus $0.09$, $0.10$, and $0.20$ for fine-tuned UniVLA).
$R^2_{\mathrm{read}}$ applies the same model and decoder to the videos that the generator produces from the true commands, and thus tests whether the commands can be read back from generated frames.
Consistent with Section~\ref{sec:conclusion}, recovering fine-grained wrist rotation and gripper commands remains substantially harder than recovering translation across all evaluated models.
In Table~\ref{cs:tab:wm-full}, Agent-State LAM matches or exceeds every compared baseline on rotation recovery ($R^2_{\mathrm{rot}}$ of $0.01$, $0.03$, and $0.14$ on IRASim, Bridge, and RT-1, versus $0.00$, $0.02$, and $0.02$ for fine-tuned UniVLA), with the clearest gain on RT-1, yet absolute rotation $R^2_{\mathrm{rot}}$ remains low for all methods.
These components may induce subtle or visually ambiguous changes that are difficult to infer from passive video alone; complementary supervision, such as language guidance, may improve their recovery in future work.

\subsection{Qualitative examples}
\label{cs:wm-examples}
\begin{figure*}[p]
\centering
\includegraphics[width=\textwidth]{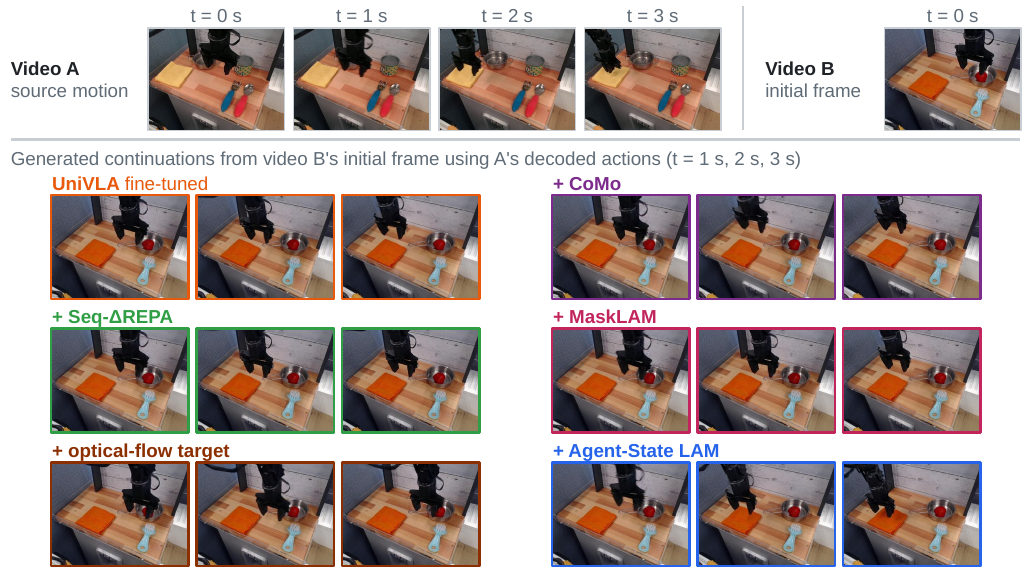}
\caption{\textbf{Cross-context motion transfer on IRASim (Example 1).} Following Figure~\ref{fig:wm}a, robot commands decoded from video A drive frozen Cosmos-Predict2 from the initial frame of video B across all six fine-tuned models. Only Agent-State LAM moves the arm down onto the cloth as in video A, whereas the baselines stop short.}
\label{cs:fig:wm-transfer}
\end{figure*}

\begin{figure*}[p]
\centering
\includegraphics[width=\textwidth]{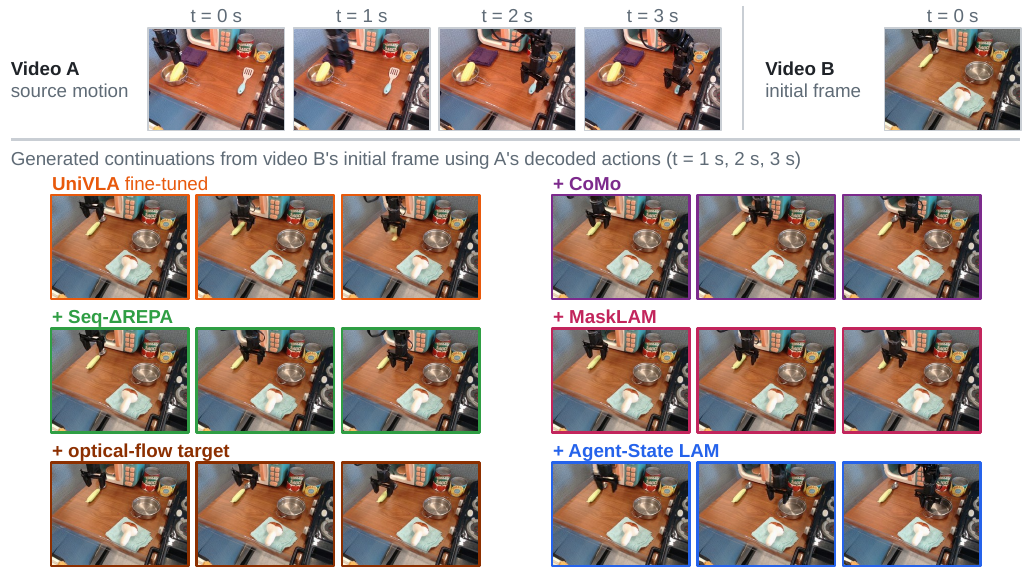}
\caption{\textbf{Cross-context motion transfer on IRASim (Example 2).} Same setting and layout as Figure~\ref{cs:fig:wm-transfer}. Only Agent-State LAM reproduces video A's rightward sweep across the table, whereas the baselines remain on the left near the corn.}
\label{cs:fig:wm-transfer-b}
\end{figure*}

\begin{figure*}[p]
\centering
\includegraphics[width=\textwidth]{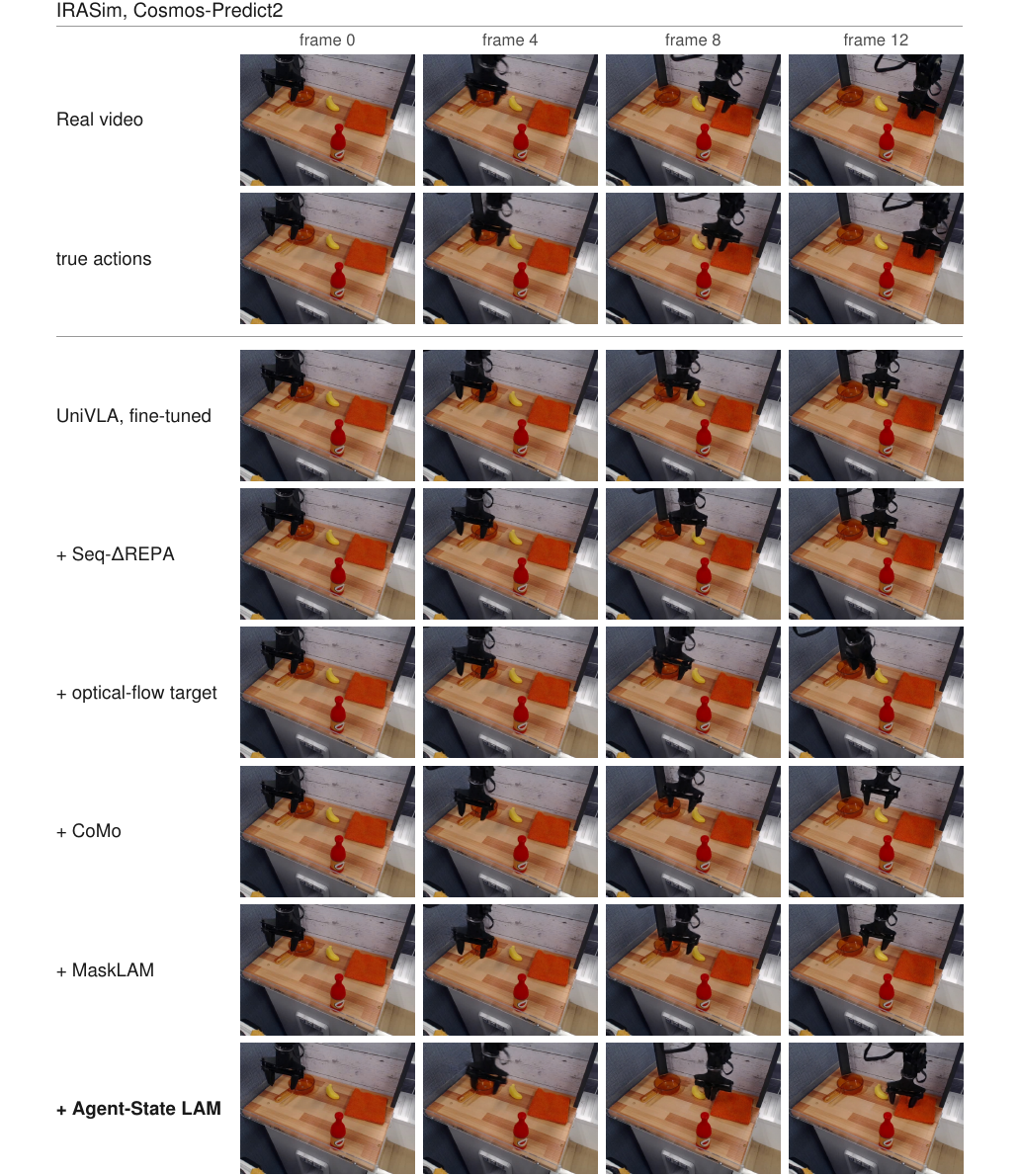}
\caption{\textbf{Same-clip world-model prediction on IRASim.} All generated rows share the clip's initial frame (frame $0$), frozen Cosmos-Predict2 generator, and random seed, differing only in their conditioning commands. True actions and Agent-State LAM move the arm from the bowl to the cloth as in the real video, whereas the baselines stop before reaching the cloth.}
\label{cs:fig:wm-examples}
\end{figure*}

\begin{figure*}[p]
\centering
\includegraphics[width=\textwidth]{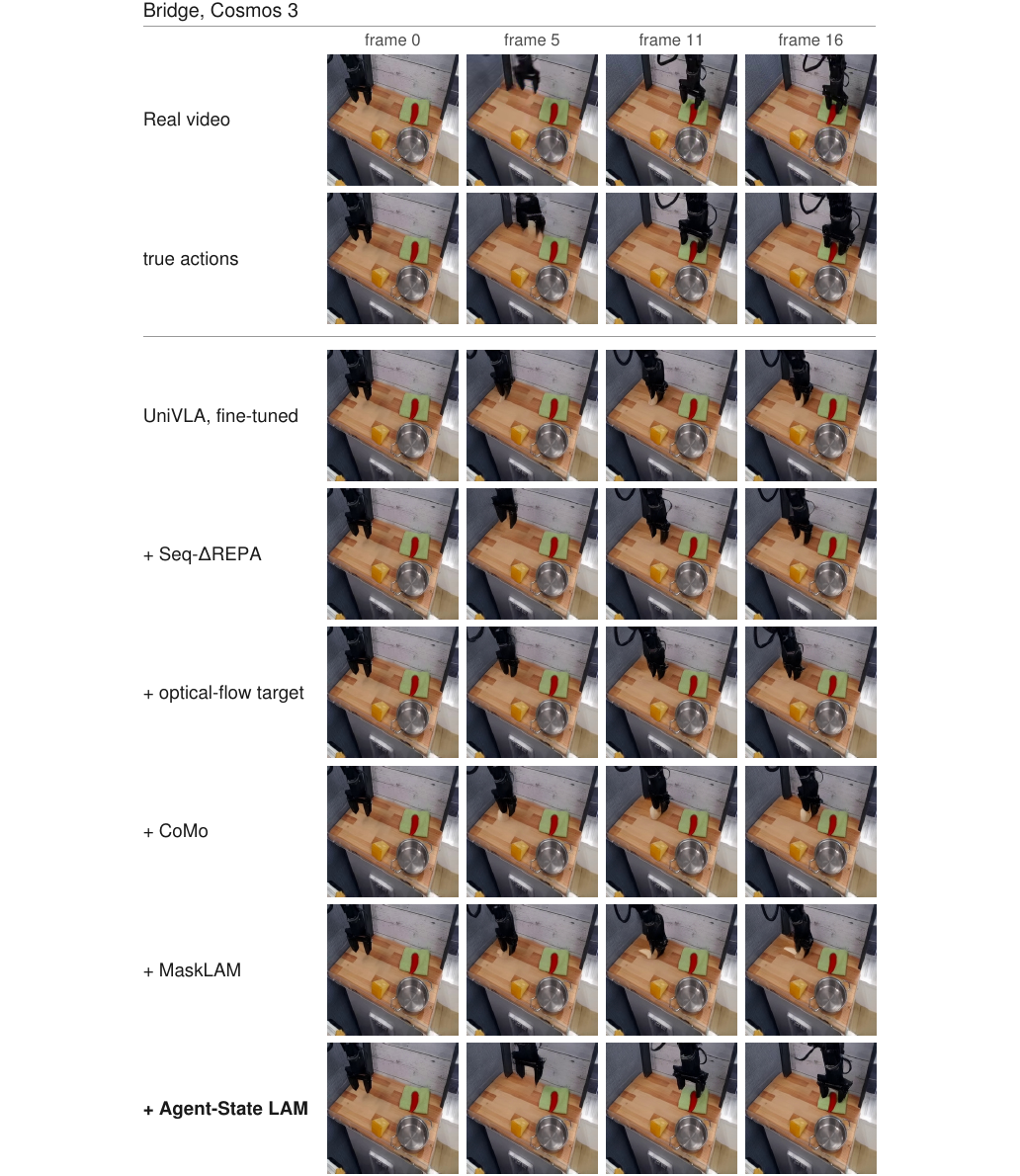}
\caption{\textbf{Same-clip world-model prediction on Bridge.} Same layout as Figure~\ref{cs:fig:wm-examples}, using frozen Cosmos~3. True actions and Agent-State LAM move the arm to the pepper on the green cloth as in the real video, whereas the baselines remain to the left of the cloth.}
\label{cs:fig:wm-examples-bridge}
\end{figure*}

\begin{figure*}[p]
\centering
\includegraphics[width=\textwidth]{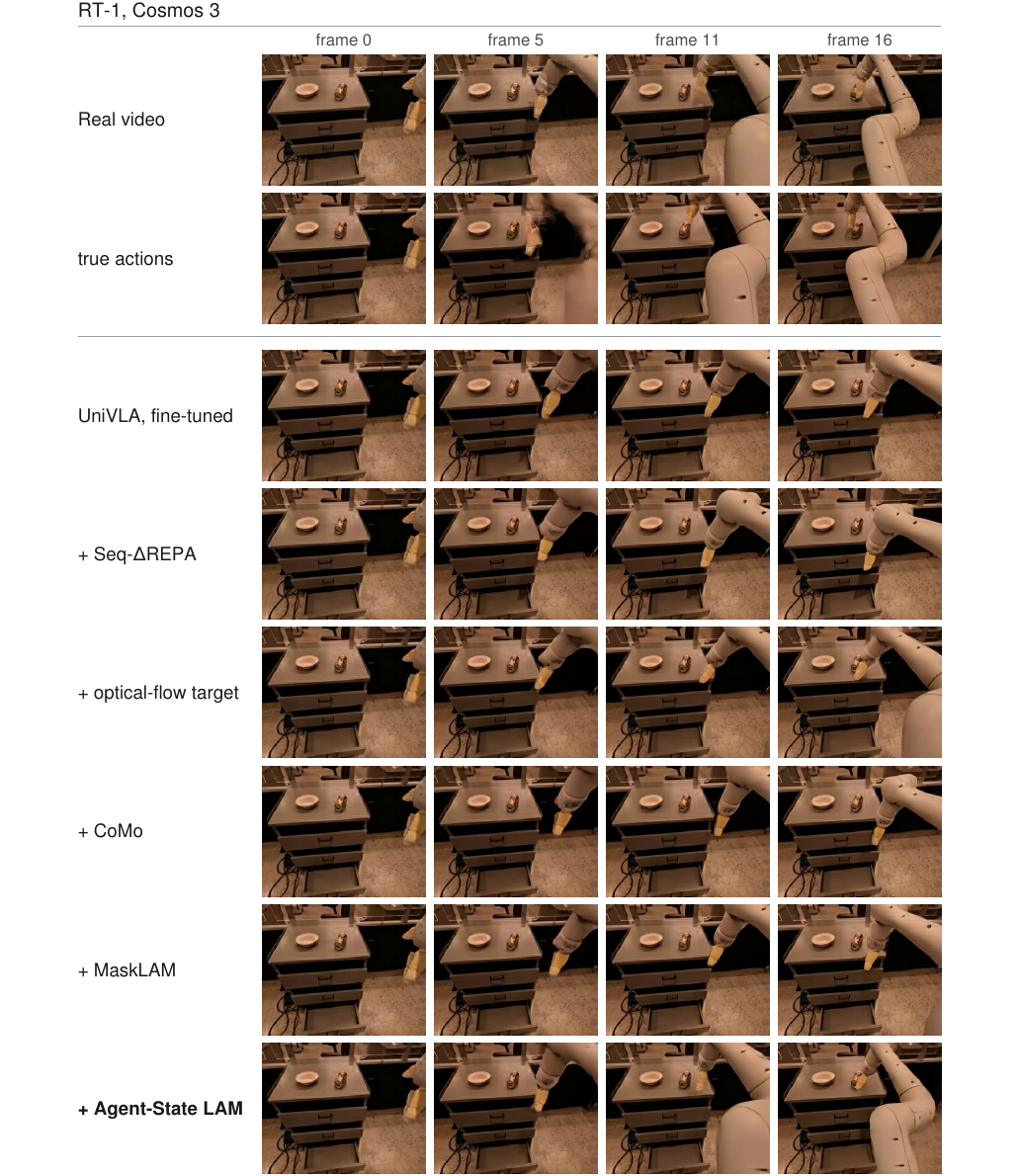}
\caption{\textbf{Same-clip world-model prediction on RT-1.} Same layout as Figure~\ref{cs:fig:wm-examples}, using frozen Cosmos~3. True actions and Agent-State LAM move the arm to the can on the counter as in the real video; the optical-flow-target variant also reaches the can, whereas the remaining baselines stop at the right edge of the counter.}
\label{cs:fig:wm-examples-rt1}
\end{figure*}

For cross-context motion transfer, Figure~\ref{fig:wm}a shows one pair of IRASim videos across three fine-tuned models.
Figures~\ref{cs:fig:wm-transfer} and~\ref{cs:fig:wm-transfer-b} show two additional pairs comparing all six fine-tuned models.
In Figure~\ref{cs:fig:wm-transfer}, the arm in video A moves down onto the cloth, and only with Agent-State LAM's commands does the generated arm in video B reach the cloth; in Figure~\ref{cs:fig:wm-transfer-b}, the arm in video A sweeps across the table to the right, and again only Agent-State LAM's commands reproduce this rightward sweep.
For same-clip prediction, Figures~\ref{cs:fig:wm-examples}--\ref{cs:fig:wm-examples-rt1} show one held-out clip per dataset under the true actions and the decoded commands of all six fine-tuned models, sharing the initial frame and random seed across the generated rows.
On IRASim (Figure~\ref{cs:fig:wm-examples}) and Bridge (Figure~\ref{cs:fig:wm-examples-bridge}), only Agent-State LAM's commands move the arm to the cloth and the pepper on the green cloth as in the real videos, whereas the baselines stop short.
On RT-1 (Figure~\ref{cs:fig:wm-examples-rt1}), both Agent-State LAM and the optical-flow-target variant move the arm to the can on the counter, consistent with Table~\ref{cs:tab:wm-full}, while the remaining models stop at the counter's right edge.

\clearpage


\end{document}